\documentclass{article} 
\usepackage{styles/iclr2027_conference,times}

\usepackage{hyperref}
\usepackage[center]{styles/VABcommands}
\usepackage{styles/VABenvironments}
\usepackage{styles/VABcitations}

\usepackage{cleveref}
\usepackage{tikz}
\usepackage{subcaption}
\usepackage{csquotes}
\usepackage{booktabs}
\usepackage{adjustbox}
\usepackage{nicefrac}
\usepackage{multirow}
\usepackage[dvipsnames]{xcolor}

\MakeOuterQuote{"}

\usetikzlibrary{positioning}
\usetikzlibrary{calc}

\title{Do Deep Ensembles Actually Capture \\* Uncertainty in Graph Neural Networks?}

\author{Pedro C. Vieira\textsuperscript{\ensuremath{1,2}}
    \And Pedro Ribeiro\textsuperscript{\ensuremath{2}}
    \And Viacheslav Borovitskiy\textsuperscript{\ensuremath{1}}
    \AND
    \normalfont
    \textsuperscript{\ensuremath{1}} University of Edinburgh
    \qquad
    \textsuperscript{\ensuremath{2}} DCC/FCUP, University of Porto
}

\bibliography{revised_references}

\hypersetup{
    colorlinks,
    linkcolor={red!50!black},
    citecolor={blue!50!black},
    urlcolor={blue!80!black}
}

\iclrpreprintcopy
\begin{document}

\maketitle

\begin{abstract}
While deep ensembles are widely considered to be the default method for uncertainty quantification in deep learning, their effectiveness for graph-structured data is often simply assumed based on successes in domains like computer vision.
We investigate standard deep ensembles specifically for message-passing graph neural networks.
Benchmarking across eight datasets representing varied tasks and complexities, we reveal that ensembles provide only marginal improvement over a single model.
These marginal gains stem primarily from averaging out optimization noise in point predictions rather than from meaningfully better uncertainty estimates.
Through an aleatoric-epistemic decomposition, we identify \emph{epistemic collapse}: independently trained networks consistently converge to overly similar predictions.
Because disagreement is the fundamental mechanism through which ensembles capture epistemic uncertainty, this lack of diversity neutralizes their key advantage.
Towards explaining this phenomenon, we derive a bound on epistemic variance in terms of the graph topology and Dirichlet energy budget, linking epistemic collapse to the smoothing bias of message passing.
Our results suggest that deep ensemble success does not seamlessly transfer to graph machine learning.
\end{abstract}

\section{Introduction}\label{sec:intro}

Uncertainty quantification is essential in high-stakes settings or automated decision-making applications~\parencite{Jaya2025-UQ-health, Zhu2020-UQ-AutoSystems, garnett2023}.
Among practical approaches to uncertainty quantification, deep ensembles (DEs)~\parencite{Lak-DE} have emerged as a widely used method.
Their appeal lies in simplicity: building a standard deep ensemble requires only the repeated training of a single architecture.
The disagreement among these independently trained models captures uncertainty that a single model cannot.
This mechanism has proven highly effective, establishing deep ensembles as the default for uncertainty quantification in deep learning, provided the added computational cost is not prohibitive~\parencite{Ovadia2019,Tran2020}.

When processing graph-structured data---such as molecules or road networks---uncertainty quantification is much less studied.
Here, message-passing neural networks, like graph convolutional (GCN)~\parencite{kipf2017semisupervised} or attention (GAT)~\parencite{velickovic2018graph} networks, have emerged as one of the dominant architecture classes.
While other classes like graph transformers~\parencite{dwivedi2020, rampavsek2022} exist, we focus exclusively on message-passing architectures, which we refer to simply as graph neural networks (GNNs) throughout this work.
Unlike standard deep learning architectures, GNN computation flow is defined by the structure of the input graph, a property that fundamentally alters the hypothesis space and how optimization trajectories explore it~\parencite{xu2018how, Rusch2023Oversmoothing}.
This structural difference can significantly affect how ensembles of such models behave, for better or worse.

Surveys and application papers often present DEs as a successful, out-of-the-box solution for uncertainty quantification in graph learning tasks~\parencite{wang2024uncertainty-survey,Heid2023}.
However, they tend not to compare the uncertainty of DEs with that of even simpler baselines: the strength of DEs in classical deep learning settings like computer vision is largely \emph{assumed} to carry over to graphs.
Works that do evaluate the uncertainty of DEs paint a mixed picture.
DEs range from hard to beat~\parencite{Trivedi2024} to barely distinguishable from a single model~\parencite{Stadler2021}, and even underperform a trivial baseline on the PEMS road network benchmark~\parencite{borovitskiy2025,mostowsky2025}---a failure we revisit in~\Cref{sec:motivation}.
In these works, DEs themselves are a baseline rather than the object of study, and, to our knowledge, no work systematically studies how good this simple and practical baseline is: how well standard DEs of GNNs capture uncertainty, and why.\footnote{Modified ensemble methods exist for specific graph tasks (discussed in \Cref{subsec:related-work}), but we focus strictly on the standard DE mechanism of repeatedly training the same architecture from different initializations.}

Motivated by these mixed signals, we provide such a study.
To this end, we systematically benchmark DEs of GNNs across multiple tasks, focusing explicitly on the quality of their uncertainty estimates.
Our results demonstrate that their behavior in graph settings falls far short of expectations inherited from classic deep learning settings, most notably computer vision, where DEs are most widely tested.
Our central findings are an empirical phenomenon we term \emph{epistemic collapse}---individual GNNs within a DE exhibit surprisingly limited diversity---and a theoretical bound on epistemic variance that links this collapse to the smoothing bias of message passing.

Our contributions are:
\1 We systematically benchmark standard DEs of GNNs across multiple classic and modern graph machine learning datasets.
\2 We empirically demonstrate the phenomenon of epistemic collapse, showing that GNNs tend to converge to highly similar functions regardless of initialization.
\3 We demonstrate that the minor improvements in the negative log-likelihood (NLL) score provided by DEs are primarily attributable to moderately improved prediction accuracy rather than better uncertainty estimates.
\4 We derive a bound on epistemic variance in terms of the graph topology and Dirichlet energy budget.
It links epistemic collapse to the smoothing bias of message passing---the same bias that helps on homophilic graphs and, in excess, causes oversmoothing---and suggests weaker collapse under heterophily, aligning with our experiments.
\0

\section{Background}\label{sec:background}

\paragraph{Graph neural networks.}
As mentioned in~\Cref{sec:intro}, we focus on \emph{message-passing GNNs} \parencite{Gilmer2017-mpnn}.
Given a graph $\mathcal{G}$ with node features $\v{x}_i$, each layer computes

\[ \label{eqn:message_passing}
\v{h}^{(\ell+1)}_i = \gamma_{\m{\Theta}_1}\del[3]{\v{h}^{(\ell)}_i, \bigoplus\nolimits_{j \in \mathcal{N}(i)} \psi_{\m{\Theta}_2}(\v{h}^{(\ell)}_i, \v{h}^{(\ell)}_j)}, && \v{h}^{(0)}_i = \v{x}_i,
\]

where $\mathcal{N}(i)$ denotes the neighbors of $i$, $\psi_{\m{\Theta}_2}$ computes messages, $\bigoplus$ is a permutation-invariant aggregation, and $\gamma_{\m{\Theta}_1}$ post-processes the result.
Architectures such as GCN \parencite{kipf2017semisupervised}, GAT \parencite{velickovic2018graph}, and GIN \cite{xu2018how} vary this template, but always follow the graph topology.

\paragraph{Uncertainty quantification and deep ensembles.}
Uncertainty quantification aims to measure not just what a model predicts, but how reliable that prediction is.
Predictive uncertainty is conceptually divided into \emph{aleatoric} uncertainty, which reflects irreducible noise in the data, and \emph{epistemic} uncertainty, which reflects the model's lack of knowledge \parencite{Kendall2017}.

Originally proposed by \textcite{Lak-DE}, deep ensembles (DEs) provide a straightforward, non-Bayesian approach to uncertainty quantification.
The method aggregates predictions from $M$ copies of the same architecture trained from different random weight initializations.
In a deep ensemble, aleatoric and epistemic uncertainty can be explicitly separated.

For regression, the $m$-th member of a DE predicts the mean $\mu_m(x)$ and variance $\sigma^2_m(x)$ of a Gaussian distribution.
The ensemble prediction is modeled as a Gaussian mixture, yielding an aggregated point prediction $\mu(x) = \frac{1}{M} \sum_{m=1}^M \mu_m(x)$.
The total predictive uncertainty is the variance of this mixture, which elegantly decomposes into the average variance and the variance of the means:
\[ \label{eqn:var_decomp}
\underbrace{\sigma^2(x)}_{\text{Total}}
=
\underbrace{\frac{1}{M} \sum\nolimits_{m=1}^M \sigma_m^2(x)}_{\text{Aleatoric } \sigma^2_{\text{ale}}(x)}
+
\underbrace{\frac{1}{M} \sum\nolimits_{m=1}^M \del{ \mu_m(x) - \mu(x) }^2}_{\text{Epistemic } \sigma^2_{\text{epi}}(x)}
.
\]

For classification, the $m$-th member of a DE outputs a probability distribution $p_m(y \given x)$ over classes.
The ensemble outputs a single distribution formed by averaging the member probabilities, $p(y \given x) = \frac{1}{M} \sum_{m=1}^M p_m(y \given x)$.
Here, total uncertainty is measured by the predictive entropy $\c{H}$, which similarly decomposes into expected data uncertainty and mutual information \parencite{pmlr-v80-depeweg18a}:
\[ \label{eqn:ent_decomp}
\underbrace{\c{H}[p(y \given x)]}_{\text{Total}}
=
\underbrace{\frac{1}{M} \sum\nolimits_{m=1}^M \c{H}[p_m(y \given x)]}_{\text{Aleatoric } \c{H}_{\text{ale}}(x)}
+
\underbrace{\del[3]{ \c{H}[p(y \given x)] - \frac{1}{M} \sum\nolimits_{m=1}^M \c{H}[p_m(y \given x)] }}_{\text{Epistemic } \c{H}_{\text{epi}}(x)}
.
\]

\subsection{Related Work}\label{subsec:related-work}

Much of the prior work on GNN ensembles departs from the standard DE recipe.
Examples include ensembling models trained on different graph views \parencite{Nguyen2025-PH}, employing varied GNN architectures \parencite{Wong2023}, or relying on bootstrapped training data \parencite{Krieg2022-HO}.
Other approaches approximate Bayesian inference by sampling from a single optimization trajectory \parencite{Rahman2024} or utilizing stochastic message passing \parencite{Lin2022}.
In specialized domains like chemistry and material sciences, ensemble methods often use customized edge-centric architectures, shallow weight sharing, or post-hoc recalibration \parencite{Varivoda2022-ensemble-almost-mpnn, Busk2021, Hirschfeld2020-dmpnn-ensembles, Vinchurkar2025-shallow-mpnn-ensemble}.

Such modifications give up the simplicity of the standard recipe.
If they were required for ensemble disagreement on graphs, this would mark a fundamental departure from classical deep learning, where random initialization alone suffices \parencite{Lak-DE,Ovadia2019}---a departure that was never systematically established.
Standard DEs of GNNs appear mainly as baselines, both for graph-specific uncertainty methods~\parencite{Stadler2021,Fuchsgruber2024energy,Fuchsgruber2024active,Trivedi2024} and in benchmarks~\parencite{Scalia2020,Bazhenov2022,Bazhenov2023-graphshift}.
Their results form the mixed picture described in~\Cref{sec:intro}, but analyzing DEs is not their aim.
We make standard DEs the object of study: we systematically evaluate what they gain in uncertainty estimation over a single GNN, which costs a fraction as much, identify epistemic collapse, and link it to the smoothing bias of message passing.

\section{A Motivating Example: The PEMS Benchmark}\label{sec:motivation}

\begin{figure}[b]
    \centering
    \begin{subfigure}[b]{0.32\textwidth}
        \centering
        \includegraphics[width=\linewidth]{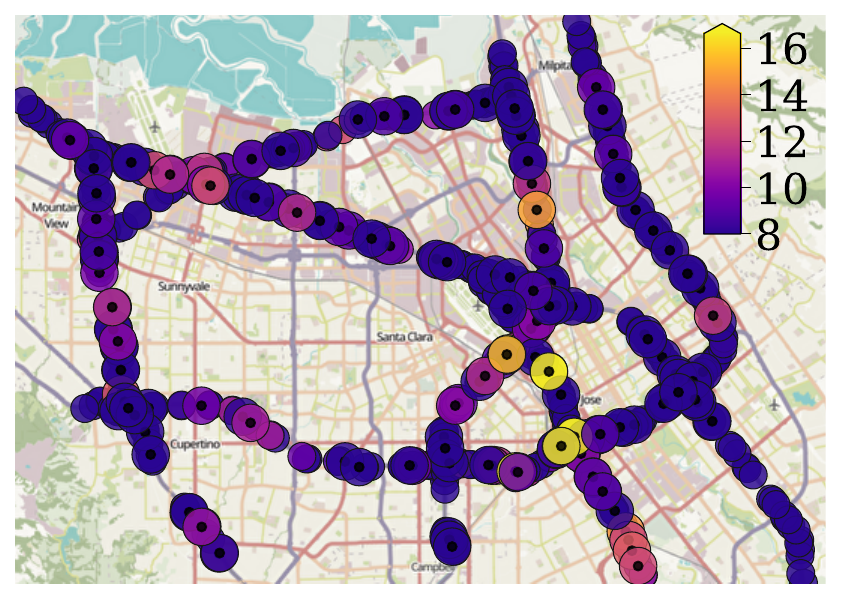}
        \caption{Uncertainty Map: GCN DE}
        \label{fig:pems_results:de_unc}
    \end{subfigure}\hfill
    \begin{subfigure}[b]{0.32\textwidth}
        \centering
        \includegraphics[width=\linewidth]{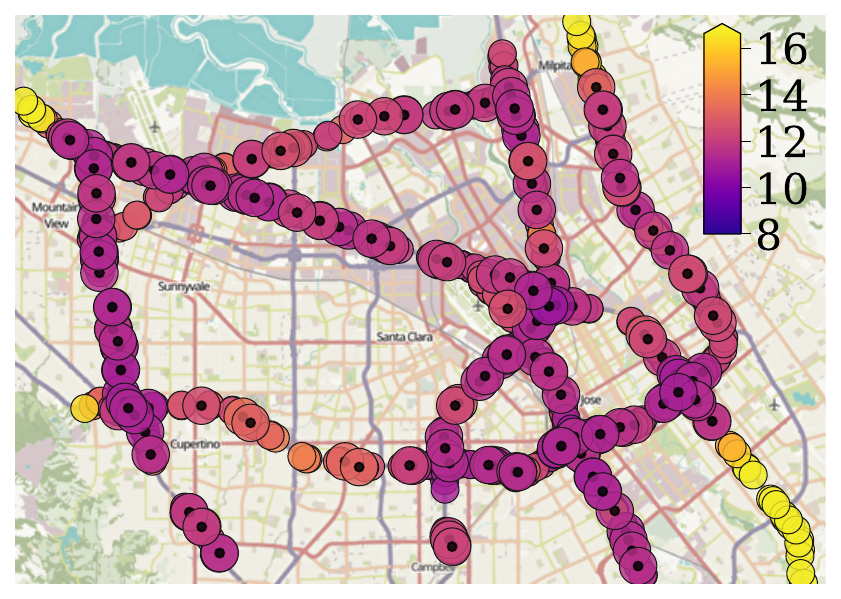}
        \caption{Uncertainty Map: Graph GP}
        \label{fig:pems_results:gp_unc}
    \end{subfigure}\hfill
    \begin{subfigure}[b]{0.33\textwidth}
        \centering
        \renewcommand{\arraystretch}{1.45}
        \begin{adjustbox}{width=\linewidth}
            \newcommand{\B}[1]{\textcolor{ForestGreen}{#1}} 
\newcommand{\W}[1]{\textcolor{BrickRed}{#1}}    

\begin{tabular}{lcc}
    \toprule
    \addlinespace
    \textbf{Method} & \textbf{NLL} ($\downarrow$) & \textbf{RMSE} ($\downarrow$) \\
    \addlinespace
    \midrule
    \addlinespace
    Trivial         & $1.37 \pm 0.00$ & $16.4 \pm 0.0$ \\
    GCN             & \W{$2.51 \pm 0.39$} & \B{$13.5 \pm 0.1$} \\
    DE          & \W{$2.39 \pm 0.11$} & \B{$13.6 \pm 0.1$} \\
    B-GCN    & \B{$1.32 \pm 0.09$} & \B{$13.4 \pm 0.1$} \\
    Graph GP        & \B{$0.98 \pm 0.02$} & \B{$11.1 \pm 0.2$} \\
    \addlinespace
    \bottomrule
\end{tabular}

        \end{adjustbox}
        \caption{Performance Metrics}
        \label{fig:pems_results}
    \end{subfigure}
    
    \vspace{0.5em}
    \caption{PEMS road network benchmark. Nodes with a black dot in the middle are train nodes. DE produces highly overconfident and uninformative uncertainty estimates (\subref{fig:pems_results:de_unc}) compared to the geometrically grounded Graph GP (\subref{fig:pems_results:gp_unc}). Table~(\subref{fig:pems_results}) confirms that DE fails at uncertainty quantification (NLL), performing significantly worse than Trivial, and only marginally improving over GCN. \textcolor{BrickRed}{Red} denotes scores worse than trivial and \textcolor{ForestGreen}{Green} better than trivial.}
    \label{fig:pems_combined}
\end{figure}

We start by revisiting the PEMS road network benchmark \parencite{borovitskiy2025}.
Originally introduced to evaluate graph Gaussian processes \parencite{borovitskiy2021}, this transductive node regression dataset tasks models with interpolating traffic speeds and quantifying the associated predictive uncertainty.
PEMS has no informative node features, so the GNNs receive the training labels as inputs, as in \textcite{Shi2021UniMP}.

We expand upon the prior evaluation of \textcite{borovitskiy2025} by following the DE recipe of \textcite{Lak-DE}.
Specifically, we equip the base GCN models with a variance head trained with the Gaussian NLL (the original benchmark uses an ensemble of prediction-only models).
We compare a single GCN, a five-model deep ensemble (DE), a Bayesian GCN (B-GCN) trained via stochastic variational inference~\parencite{Louizos2017,Wen2018}, and a Matérn Gaussian process on the graph (Graph GP) \cite{borovitskiy2021}.
As a sanity check, we include a \emph{Trivial} baseline that predicts a constant Gaussian distribution $\f{N}(\mu_{\text{train}}, \sigma^2_{\text{train}})$ fitted to the training labels.
For more details on the setup, see~\Cref{sec:appx-extra-details-exp}.

The results, shown in \Cref{fig:pems_combined}, reveal a failure of the deep ensemble.
It improves RMSE over the Trivial baseline, but its NLL is much worse.
The single GCN is already worse than Trivial, because its variance head is overconfident.
A DE could repair this, but it does not: its NLL barely improves.
Visually (\Cref{fig:pems_results:gp_unc,fig:pems_results:de_unc}), the Graph GP produces an uncertainty map that sensibly increases away from training data.
In contrast, the DE's uncertainty is uniformly low and sometimes spikes at training nodes.
The Bayesian GCN, whose hidden dimension we had to reduce to train it (\Cref{sec:appx-extra-details-exp}), beats the Trivial baseline, so the failure here is specific to deep ensembles of GNNs, not to GNNs alone.
Finally, the Graph GP performs best, as expected: the signal is a smooth function on the graph, the setting graph GPs were designed for, and there are no rich node features, whose processing is where GNNs normally have the edge.

This motivating example raises a critical question: \emph{Is the failure on PEMS a symptom of a systematic weakness of deep ensembles of graph neural networks, and if so, what causes this weakness?}

\section{Benchmarking Ensembles of Graph Neural Networks}\label{sec:more-experiments}

To test whether the failure on \texttt{PEMS} points to a systematic weakness, we conduct a comprehensive evaluation of deep ensembles of GNNs across a diverse range of graph learning tasks.

\begin{table}[b]
    \vspace{-0.5\baselineskip}
    \centering
    \renewcommand{\arraystretch}{1.15}
    \caption{Absolute performance metrics for a DE of five models. Values represent NLL, ECE (for regression stands for miscalibration area), and point estimation metric (PEM: RMSE for regression and accuracy for classification), showing the mean and standard deviation over 10 independent runs. The ECE of the trivial baseline is not taken into account when highlighting the best score. Trivial baseline has low ECE by construction but is uninformative: it predicts the same regardless of input.}
    \vspace{0.5\baselineskip}
    \label{tab:main_results_table}
    \begin{adjustbox}{max width=\linewidth}
    \begin{tabular}{cl ccc @{\hspace{3em}} ccc}
    \toprule
    & & \multicolumn{3}{c}{\textbf{Classification}} & \multicolumn{3}{c}{\textbf{Regression}} \\
    \cmidrule(r){3-5} \cmidrule(l){6-8}
    & & \textbf{CTSR} & \textbf{TLK2} & \textbf{MHIV} & \textbf{ARTNV} & \textbf{CHAM} & \textbf{QM9-5\%} \\
    \midrule

    \multirow{3}{*}{\rotatebox[origin=c]{90}{\textbf{Trivial}}}
    & NLL & $1.792 \pm 0.000$ & $0.525 \pm 0.000$ & $0.141 \pm 0.000$ & $1.425 \pm 0.000$ & $1.415 \pm 0.000$ & $1.404 \pm 0.000$ \\
    & ECE & $0.069 \pm 0.000$ & $0.000 \pm 0.000$ & $0.006 \pm 0.000$ & $0.015 \pm 0.000$ & $0.072 \pm 0.000$ & $0.070 \pm 0.000$ \\
    & PEM & $0.077 \pm 0.000$ & $0.782 \pm 0.000$ & $0.968 \pm 0.000$ & $1.715 \pm 0.000$ & $2.160 \pm 0.000$ & $1.268 \pm 0.000$ \\

    \midrule

    \multirow{3}{*}{\rotatebox[origin=c]{90}{\textbf{GNN}}}
    & NLL & $\boldsymbol{0.929 \pm 0.003}$ & $0.371 \pm 0.002$ & $0.115 \pm 0.001$ & $0.993 \pm 0.005$  & $0.868 \pm 0.066$  & $0.833 \pm 0.021$ \\
    & ECE & $0.142 \pm 0.002$ &  $0.079 \pm 0.000$ & $0.057 \pm 0.004$ & $0.033 \pm 0.017$ & $\boldsymbol{0.137 \pm 0.054}$ & $\boldsymbol{0.568 \pm 0.014}$ \\
    & PEM & $0.710 \pm 0.004$ &  $0.854 \pm 0.005$ & $0.970 \pm 0.001$ & $1.113 \pm 0.006$  & $1.215 \pm 0.044$  & $0.384 \pm 0.014$ \\

    \midrule

    \multirow{3}{*}{\rotatebox[origin=c]{90}{\textbf{DE}}}
    & NLL & $\boldsymbol{0.929 \pm 0.001}$ & $\boldsymbol{0.367 \pm 0.001}$ & $\boldsymbol{0.112 \pm 0.001}$ & $\boldsymbol{0.958 \pm 0.004}$ & $\boldsymbol{0.749 \pm 0.026}$ & $\boldsymbol{0.803 \pm 0.003}$ \\
    & ECE & $\boldsymbol{0.141 \pm 0.002}$ & $\boldsymbol{0.077 \pm 0.005}$ & $\boldsymbol{0.054 \pm 0.003}$ & $\boldsymbol{0.009 \pm 0.002}$ & $0.266 \pm 0.027$ & $0.595 \pm 0.004$ \\
    & PEM & $\boldsymbol{0.711 \pm 0.000}$ & $\boldsymbol{0.856 \pm 0.002}$ & $\boldsymbol{0.972 \pm 0.001}$ & $\boldsymbol{1.086 \pm 0.004}$ & $\boldsymbol{1.132 \pm 0.022}$ & $\boldsymbol{0.362 \pm 0.005}$ \\

    \bottomrule
    \end{tabular}
    \end{adjustbox}
\end{table}


\paragraph{Datasets.}
In total, we evaluate on eight diverse graph datasets spanning node- and graph-level classification and regression.
Having established the failure on the PEMS (\texttt{PEMS}) road network in \Cref{sec:motivation} and deferring the simplest Cora (\texttt{CORA})~\parencite{Yang2016CoraCite} to \Cref{sec:appx-extra-plots}, we focus here on the remaining six datasets.
For node classification, we include the classic CiteSeer (\texttt{CTSR})~\parencite{Yang2016CoraCite} and a newer, larger-scale Tolokers2 (\texttt{TLK2}) \cite{bazhenov2025graphland}.
For node regression, we use Artnetviews (\texttt{ARTNV}) \cite{bazhenov2025graphland} and Chameleon (\texttt{CHAM}) \cite{Rozemberczki2021-cham}, which both offer modern, structurally challenging evaluation settings.
\texttt{CHAM}, specifically, exhibits heterophily~\parencite{Luan2024-cham-hetero} in a benign form still amenable to message passing.
Finally, for graph-level classification we use MolHIV (\texttt{MHIV}) from the Open Graph Benchmark~\parencite{hu2020ogb}, and for graph-level regression, we use the classic QM9 dataset~\parencite{Wu2018-QM9}.
\texttt{MHIV} is highly imbalanced (about 3.5\% positives), so accuracy barely separates any model from the Trivial baseline (NLL remains informative).
We subsample QM9 to 5\% of its original size (denoted \texttt{QM9-5\%}) and restrict to the HOMO-LUMO gap prediction task, one of the most difficult in the dataset.
We stratify this subsampling by the target variable to preserve its overall distribution.
We deliberately implement this subsampling to test a data-scarce regime, a setting where the quality of uncertainty estimates is especially critical.
Together, these datasets capture diverse graph learning settings.

\paragraph{Models.}
We select architectures that represent standard practice for each task.
For \texttt{CTSR}, \texttt{ARTNV}, and \texttt{QM9-5\%}, we employ standard GCNs~\parencite{kipf2017semisupervised}.
On \texttt{ARTNV}, the GCN is the established state-of-the-art~\parencite{bazhenov2025graphland}.
While attention-based models can yield marginal improvements on \texttt{CTSR}, GCNs are often preferred for their superior training stability and simplicity, hence our choice.
For \texttt{QM9-5\%}, although top-performing methods utilize 3D spatial coordinates, GCNs perform comparably to other purely topological GNNs \cite{Wu2018-QM9,fung2021} and provide a clean, representative baseline.
Conversely, we use GATv2~\parencite{brody2022gatv2} for \texttt{CHAM} and \texttt{TLK2}, where evidence favors attention; on \texttt{TLK2}, it is the state-of-the-art~\parencite{bazhenov2025graphland}.
For \texttt{MHIV}, we use GIN, the strongest standard message-passing baseline on the OGB leaderboard~\parencite{hu2020ogb}.
For regression tasks, we equip each architecture with an additional variance head to capture aleatoric uncertainty, jointly optimizing the predictive mean and variance by maximizing the likelihood.
All models are optimized to achieve performance levels commensurate with the established literature.
\Cref{sec:appx-extra-details-exp} provides more details.

\paragraph{Metrics.}
Our primary metric for jointly evaluating predictive performance and uncertainty quality is negative log-likelihood (NLL).
It is a strictly proper scoring rule \parencite{Gneiting2007}: in expectation it is minimized only by the true predictive distribution.
It is also the standard metric in the DE literature \parencite{Lak-DE,Ovadia2019}.
For finer-grained insights, we evaluate separate metrics that isolate predictive accuracy from uncertainty calibration.
Specifically, we report the root mean squared error (RMSE) for regression and accuracy for classification.
To assess calibration, we compute the expected calibration error (ECE) \cite{Guo2017} for classification and its regression analog, expected coverage error (ECE; termed \emph{miscalibration area} in \textcite{Tran2020}, which we multiply by 2 to match the expected calibration error's range).
We report the mean and standard deviation of these metrics across ten independent runs.

\begin{figure}[t]
    \centering
    \begin{subfigure}[b]{0.56\textwidth}
        \centering
        \includegraphics[height=2.75cm]{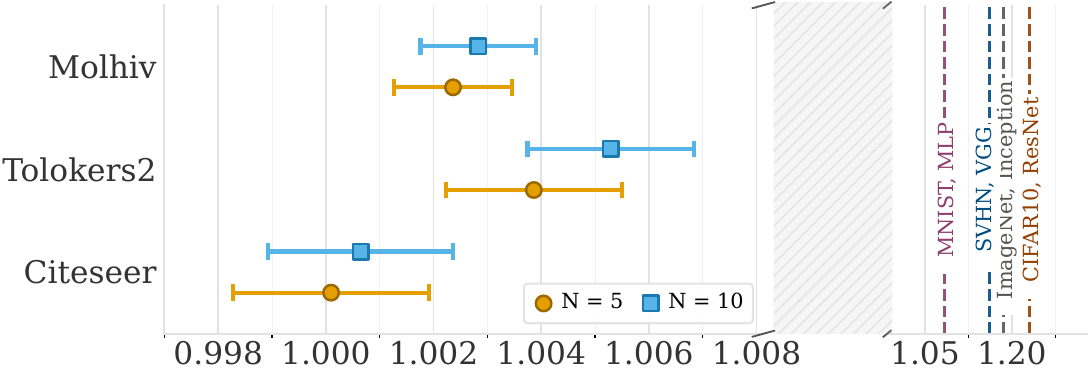}
        \caption{Classification datasets}
        \label{fig:main_results_plot:classification}
    \end{subfigure}%
    \hfill
    \begin{subfigure}[b]{0.42\textwidth}
        \centering
        \includegraphics[height=2.75cm]{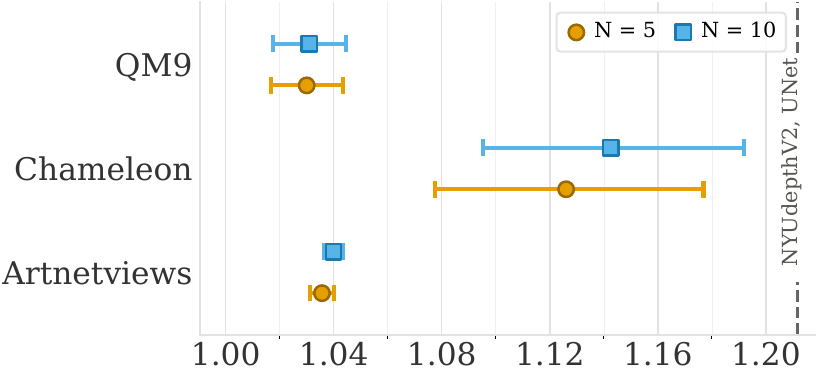}
        \caption{Regression datasets}
        \label{fig:main_results_plot:regression}
    \end{subfigure}
    
    \vspace{0.5em}
    \caption{Likelihood ratios \( \exp(\text{NLL}_{\text{GNN}} - \text{NLL}_{\text{DE}}) \) and 95\% Gaussian confidence intervals. This quantifies the relative predictive likelihood improvement of DE over GNN for both classification and regression datasets. Baselines indicate typical improvements observed in foundational DE literature.}
    \label{fig:main_results_plot}
    \vspace{-0.3cm}
\end{figure}

\paragraph{Results.}\looseness=-1
In contrast to the pathological failure on \texttt{PEMS}, \Cref{tab:main_results_table} shows that deep ensembles generally outperform the Trivial baseline across all other datasets.
They improve both point estimation and NLL compared to a single GNN in most cases, though the resulting gains appear modest and calibration (ECE) sometimes degrades, as seen on \texttt{CHAM} and \texttt{QM9-5\%}.
To properly analyze this behavior, we must ask: how do these improvements actually compare to the established literature?

To answer this question and quantify the performance gap, we compute the likelihood ratios: \( \exp\del{\text{NLL}_{\text{GNN}} - \text{NLL}_{\text{DE}}} \).
They provide an interpretable measure of the relative gain of an ensemble over a single model, representing the multiplicative factor by which the ensemble increases the likelihood of the test labels.\footnote{We avoid reporting relative percentage changes in NLL. Since log-likelihood lacks an absolute zero and can be negative, especially in continuous settings, direct ratios are mathematically flawed. We instead report the exponentiated difference, providing a sound and intuitive likelihood ratio.}
\Cref{fig:main_results_plot} compares these ratios against historical improvements derived from foundational deep ensemble literature predominantly anchored in computer vision~\parencite{Lak-DE,Bui2024-density-regression,Liu2020-spectral-gp}.

For classification tasks, the contrast is drastic: while classic computer vision benchmarks report ensembles assigning roughly 20\% more likelihood to test points than a single model, our graph ensembles yield improvements below 1\%.
In regression, the reference computer vision baseline similarly shows a $\sim$20\% improvement, whereas our graph ensembles typically assign only about 5\% more likelihood to the test data.
Even the most favorable case, \texttt{CHAM}, peaks at roughly 14\% improvement, which is not only well below the standard computer vision baseline but also exhibits a wide confidence interval, meaning the gains of a DE are highly unstable.

Crucially, \Cref{tab:main_results_table} reveals that improvements in NLL are consistently coupled with improvements in accuracy or RMSE.
Because NLL is a composite metric, these likelihood gains might be driven entirely by slightly more accurate point estimates rather than genuinely improved uncertainty calibration.
To determine whether deep ensembles on graphs actually fulfill their primary purpose---improving uncertainty quantification---we must further analyze these mechanisms in isolation.

\paragraph{Larger ensembles and distribution shift.}
While we report results for ensembles of five models here, we attest that these limitations are persistent.
As detailed in~\Cref{sec:appx-extra-plots}, scaling the ensembles to ten models yields virtually identical trends.
Additionally, evaluating deep ensembles under covariate shift on purpose-built splits for \texttt{ARTNV} and \texttt{TLK2} reveals that the relative improvements over a single GNN remain just as marginal, if not worse, than in the in-distribution setting (\Cref{sec:appx-data-shift}).

\section{How Deep Ensembles Improve Uncertainty}\label{sec:analysis}

\begin{figure}[t]
    \centering
    \begin{subfigure}[b]{0.48\textwidth}
        \centering
        \includegraphics[width=\linewidth]{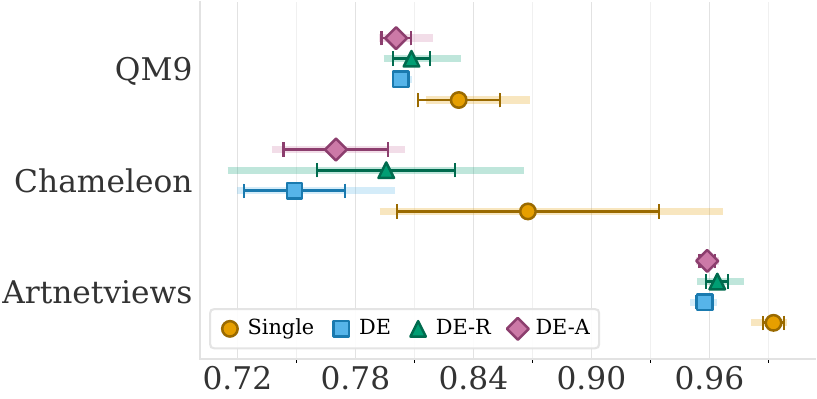}
        \caption{Regression: Comparing the NLL}
        \label{fig:disentangle:regression}
    \end{subfigure}
    \hfill
    \begin{subfigure}[b]{0.48\textwidth}
        \centering
        \includegraphics[width=\linewidth]{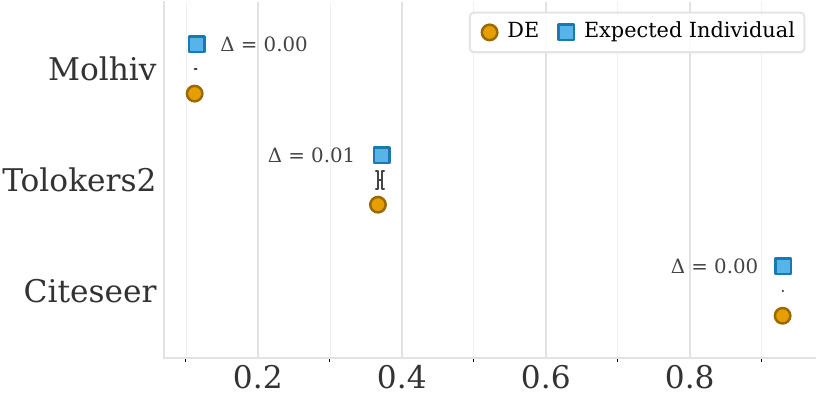}
        \caption{Classification: Jensen's Gap}
        \label{fig:disentangle:classification}
    \end{subfigure}
    
    \vspace{0.5em}
    \caption{Deconstructing the source of NLL improvements.
    (\subref{fig:disentangle:regression}) compares the NLL of a GNN, a DE, and two additional baselines (DE-R and DE-A) that isolate the point prediction improvement effect.
    (\subref{fig:disentangle:classification}) visualizes the nonnegative gap between the expected individual NLL and the ensemble's NLL (Jensen's gap). The marker represents mean, thin bars represent $\pm\sigma$ and transparent ones min--max.}
    \label{fig:disentangling}
    \vspace{-0.15cm}
\end{figure}

While \Cref{sec:more-experiments} shows that deep ensembles somewhat reduce NLL, we must disentangle whether these gains reflect genuinely better uncertainty quantification or merely more accurate point predictions.
To this end, in \Cref{sec:analysis:deconstructing}, we deconstruct the improvements of deep ensembles, revealing that the modest gains in NLL are driven almost entirely by variance reduction and predictive accuracy rather than better uncertainty estimates.
Further, in \Cref{sec:analysis:diversity}, we investigate \emph{why} this occurs by analyzing ensemble diversity.
We identify a systemic phenomenon we term \emph{epistemic collapse}, wherein individual GNNs in an ensemble fail to explore diverse functional hypotheses.

\subsection{Deconstructing the NLL Gains}\label{sec:analysis:deconstructing}

Because regression and classification parameterize uncertainty differently, we require distinct methodologies to trace the source of the NLL improvements.
For regression, we can explicitly decouple the predicted mean from the predicted variance to isolate point prediction gains.
For classification, where point estimates and uncertainty are inherently entangled in the predicted probability distribution, we instead analyze the distributional benefit of ensembling via Jensen's inequality gap.

\textbf{Regression.}
To isolate the contribution of the ensemble's point prediction, we evaluate two constructed baselines alongside a standard single GNN and a full DE.
The first baseline pairs the DE's mean prediction---the average of the individual means---with the variance of a randomly selected single model (DE-R).
The second baseline pairs the DE's mean prediction with only the aleatoric variance (DE-A), which is the average of the individual variances (cf.~\Cref{eqn:var_decomp}).

\Cref{fig:disentangle:regression} displays the NLL performance for the GNN, DE, DE-R, and DE-A models.
On \texttt{ARTNV} and \texttt{QM9-5\%}, all models utilizing the DE mean (DE, DE-R, DE-A) achieve nearly identical NLL improvements over the single model.
This is somewhat less so for \texttt{CHAM}, as \Cref{sec:theory} later on suggests is plausible under heterophily.
Thus, the already modest NLL gains are driven mainly by the ensembled mean providing a smoothed, slightly more accurate point prediction, rather than the ensemble supplying better-calibrated uncertainty.
This explains how in \Cref{tab:main_results_table} DEs can improve NLL on datasets like \texttt{CHAM} and \texttt{QM9-5\%} while simultaneously worsening the calibration error (ECE).
Furthermore, the distribution of NLL scores in \Cref{fig:disentangle:regression} reveals another key insight.
We observe that the best-performing single models can sometimes reach NLL scores comparable to those of the full ensemble or get extremely close.\footnote{Because data splits are fixed across runs, the variance stems entirely from random initialization/batching.}
This suggests that ensembling here acts primarily as a variance reduction technique over optimization noise, stabilizing the performance of a single model.

\textbf{Classification.}
Unlike regression, classification inherently entangles point estimates and confidence scores in a single predicted probability distribution.
To understand if the ensemble genuinely improves this distribution---for instance, by tempering overconfident predictions---we analyze the gap between the expected NLL of an individual model and the NLL of the ensemble.
The ensemble averages the members' class probabilities, so by Jensen's inequality its NLL is at most the average NLL of its members.
The difference (Jensen's gap) is zero exactly when all members assign the same probability to the true class, and it grows with their disagreement.

As shown in \Cref{fig:disentangle:classification}, this gap is remarkably small across all tested classification datasets.
This near-zero gap implies that the individual models are outputting practically identical probability distributions.
Because the ensembled distributions are virtually indistinguishable from the individual ones, the ensemble provides no fundamental enhancement to the predictive capabilities of the model.
Instead we observe that ensembling acts as a variance reduction mechanism over optimization noise.
The variance in performance across random initializations heavily dominates the actual ensembling effect, meaning the best-performing individual models in our runs sometimes nearly match or even improve the NLL of the full ensemble.

\subsection{Diversity and Epistemic Collapse}\label{sec:analysis:diversity}

\begin{figure}[t]
    \centering
    \begin{subfigure}[b]{0.48\textwidth}
        \centering
        \includegraphics[width=\linewidth]{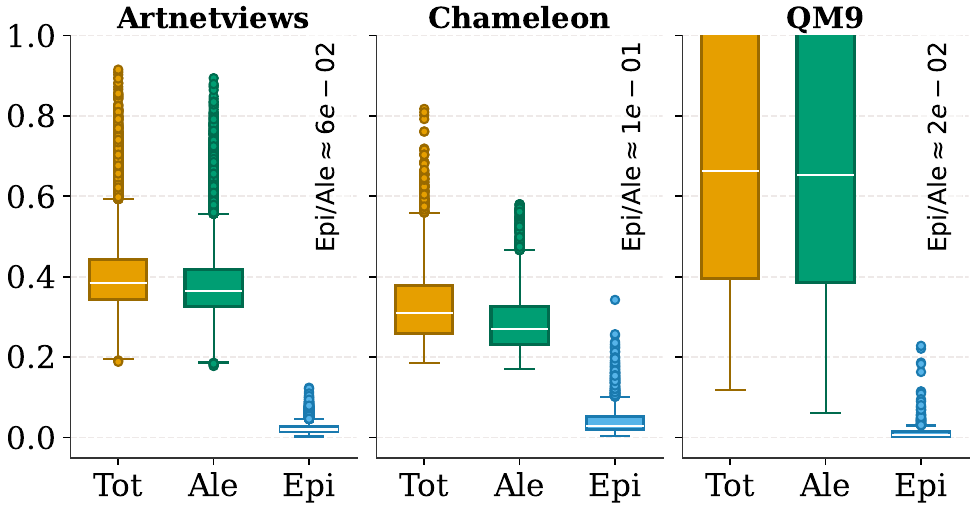}
        \caption{Uncertainty breakdown for regression.}
        \label{fig:uq-decomposition:reg}
    \end{subfigure}
    \hfill
    \begin{subfigure}[b]{0.48\textwidth}
        \centering
        \includegraphics[width=\linewidth]{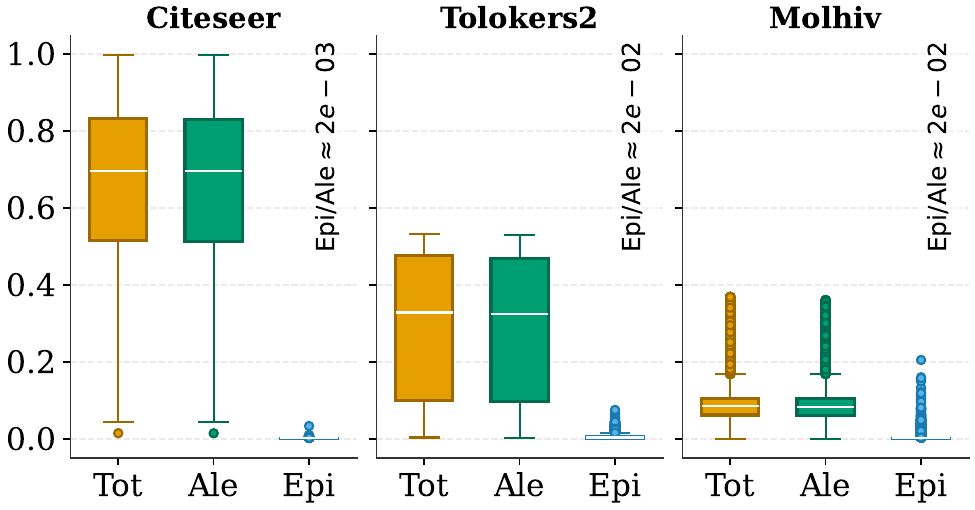}
        \caption{Uncertainty breakdown for classification.}
        \label{fig:uq-decomposition:class}
    \end{subfigure}
    
    \vspace{0.5em}
    \caption{Comparison of epistemic, aleatoric, and total uncertainty across datasets, demonstrating the \emph{epistemic collapse}.
    For both regression and classification, the epistemic part of uncertainty collapses to near-zero values, while aleatoric uncertainty remains relatively large.}
    \label{fig:uq-decomposition}
    \vspace{-0.15cm}
\end{figure}

In \Cref{sec:analysis:deconstructing}, we established that DEs of GNNs act primarily as variance-reducing smoothers.
To understand why this happens, we must consider the fundamental mechanism deep ensembles use to quantify uncertainty: the disagreement among their constituent models \parencite{Lak-DE,Abe2022-DE-Necessary,Dietterich2000}.
This disagreement is the epistemic component of uncertainty (\Cref{eqn:var_decomp,eqn:ent_decomp}), the main contribution of a DE over a single model.
If the members barely disagree, the ensemble reduces to a smoother.
We therefore investigate whether GNN ensembles suffer from what we term \emph{epistemic collapse}: a systemic failure of independently trained models to explore diverse functional hypotheses.

\Cref{fig:uq-decomposition} breaks the predictive uncertainty into its aleatoric and epistemic parts.
Across all tasks, the aleatoric part dominates.
In regression, the epistemic variance is dwarfed by the aleatoric variance; in classification, the epistemic entropy normalized by $\log C$ is close to zero, rarely exceeding~0.1.
This deviates sharply from classic deep learning settings.
For instance, \textcite{Kendall2017} report an epistemic-to-aleatoric ratio of roughly 5 for in-distribution regression in computer vision, while our ratios are well below 1.
In classification, epistemic collapse has been reported mainly for highly over-parameterized models or massive ensembles \parencite{Fellaji2024-EpistemicHole,kirsch2025implicit}; with similar normalization, we observe even smaller values across standard architectures and datasets.
Importantly, the members agree even where more epistemic uncertainty is warranted: under covariate shift, where errors grow, the DE's NLL gain stays below 3\% (\Cref{sec:appx-data-shift}), and on \texttt{PEMS}, the Graph GP shows that much more epistemic uncertainty pays off in NLL.
Ultimately, because GNNs converge to highly similar functions, they neutralize the exact mechanism deep ensembles rely on to improve uncertainty quantification.

The results thus point towards epistemic collapse: despite independent initializations, the members converge to very similar predictions.
Do they also converge to similar weights?
If so, averaging them into a single network (a model soup, \cite{wortsman2022}) would perform at least as well as a single member.
Instead, it performs generally worse (\Cref{sec:appx-model-soup}), as expected from the permutation symmetries of neural networks \parencite{entezari2022,ainsworth2023}.
This suggests the collapse is thus functional: the members differ in weights yet implement nearly the same function.

\section{Why Ensemble Members Agree: A Smoothness Bound}\label{sec:theory}

\Cref{sec:analysis:diversity} showed that the members of DEs of GNNs barely disagree.
We now take a first step towards explaining why, concentrating specifically on node-level tasks.
We show that if the members' predictions are smooth on the graph and fit the training nodes, their disagreement is bounded by this smoothness and by how well the graph connects the other nodes to the training nodes.

\begin{figure}[t]
  \centering
  \begin{subfigure}[b]{0.328\textwidth}
    \centering
    \includegraphics[width=\linewidth]{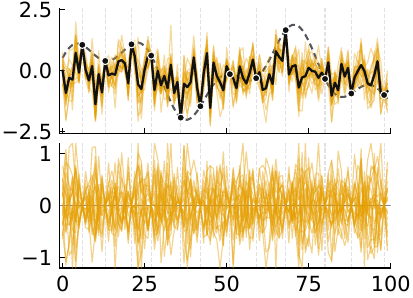}
    \vspace{-5pt} 
    \subcaption{2-layer MLP}
  \end{subfigure}%
  \hspace{0.005\textwidth}
  \begin{subfigure}[b]{0.328\textwidth}
    \centering
    \includegraphics[width=\linewidth]{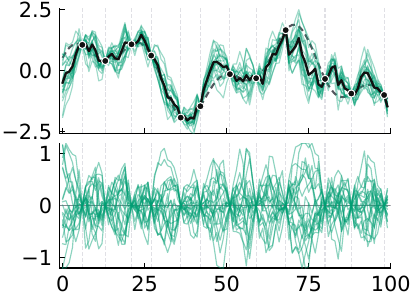}
    \vspace{-5pt}
    \subcaption{1-layer GCN}
  \end{subfigure}%
  \hspace{0.005\textwidth}%
  \begin{subfigure}[b]{0.328\textwidth}
    \centering
    \includegraphics[width=\linewidth]{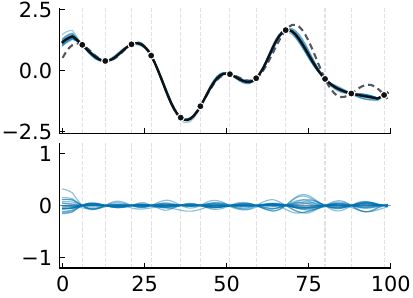}
    \vspace{-5pt}
    \subcaption{4-layer GCN}
  \end{subfigure}

  \vspace{-5pt} 
  \caption{
    Smoothing limits the disagreement.
    Ensembles trained on a path graph with 100 nodes; the $x$-axis is the position along the path.
    Top: target (dashed), anchors (dots), member predictions (colored) and ensemble mean (solid black).
    Bottom: member deviations from the ensemble mean.
  }
  \label{fig:line-graph-example}
  \vspace{-0.15cm}
\end{figure}

\paragraph{Intuition.}
Consider the deviation of a DE member's predictions from the ensemble mean as a signal on the nodes of the graph.
At the training nodes, which we call \emph{anchors}, this deviation is close to zero, as members fit the labels.
If the members' predictions change little between neighboring nodes, so do their deviations.
A signal that is zero at the anchors and changes little between neighboring nodes cannot grow large, unless some nodes are poorly connected to the anchors.
\Cref{fig:line-graph-example} illustrates this on a path graph, for an MLP and GCNs of different depth.
All models fit the anchors perfectly, and the deviations vanish at them.
An MLP is oblivious to the graph structure, so between the anchors its members disagree freely from node to node.
The deviations of GCN members are smooth, and the deeper GCN, which smooths more, barely disagrees.
We now make this argument precise, measuring how much predictions change between neighboring nodes by the \emph{Dirichlet energy} \parencite{Rusch2023Oversmoothing}.

\paragraph{Setup.}
Let $G$ be a graph with $n$ nodes and a graph Laplacian $\m{L}$, and let $S$ be the set of anchors (training nodes with labels $y_s$), such that every connected component of $G$ contains at least one anchor.
Let $\v{f}_w \in \R^n$ be the predictions $f_w(v)$ of a model with trained weights $w$ at all nodes $v$.
Randomness in initialization and training makes $w$, and hence $\v{f}_w$, random: DE members are merely independent draws of $\v{f}_w$.
Denoting the Dirichlet energy of $\v{z} \in \R^n$ by $\c{E}(\v{z}) = \v{z}^\top \m{L}\, \v{z}$, we make two assumptions.
\textbf{Assumption 1:} predictions are smooth in expectation, $\E_w[\c{E}(\v{f}_w)] \leq \rho$, where we call $\rho$ the \emph{energy budget}.
This reflects the smoothing bias of message passing \parencite{Ma2022Homophily,Luan2024Heterophilic}, which we discuss below.
\textbf{Assumption 2:} predictions fit the anchors, $\E_w[(f_w(s) - y_s)^2] \leq \varepsilon^2$ for all $s \in S$.

\begin{theorem*}
Under Assumptions 1 \& 2, the total epistemic variance $V_{\text{epi}} = \sum_{v} \Var_w(f_w(v))$ satisfies
\[ \label{eq:epistemic-bound}
V_{\text{epi}}
\leq
\inf_{\alpha > 0} \frac{\rho - \c{E}_{\min} + \alpha |S| \varepsilon^2}{\lambda_\alpha},
\]
where $\c{E}_{\min} = \min\{\c{E}(\v{z}) : \abs{z_s - y_s} \leq \varepsilon \text{ for all } s \in S\}$ is the \emph{minimal fit energy}, the least energy needed to fit the anchors within $\varepsilon$ error, and $\lambda_\alpha = \min_{\norm{\v{x}} = 1} \del[1]{\c{E}(\v{x}) + \alpha \sum_{s \in S} x_s^2}$ is the \emph{anchored spectral gap}, which is small when some nodes are poorly connected to the anchors.
\end{theorem*}

The proof (\Cref{sec:appx-theory}) roughly follows the intuition above, combining Assumptions 1 and 2 with a Poincaré-type inequality.
If predictions fit the anchors exactly ($\varepsilon = 0$), letting $\alpha \to \infty$ gives $V_{\text{epi}} \leq (\rho - \c{E}_{\min}) / \lambda_U$, where $\lambda_U$ is the smallest eigenvalue of $\m{L}$ restricted to the non-anchor nodes.
The bound also covers classification, where a model predicts a vector of class probabilities at each node.
The theorem holds for such vector-valued predictions, with squares replaced by squared norms and one-hot labels as targets.
Assumption 2 then bounds the model's Brier score at each anchor.
The bound thus limits how much the DE members' class probabilities can disagree, and therefore epistemic uncertainty, since close probability vectors have close entropies.

The \emph{spare energy} $\rho - \c{E}_{\min}$ is the part of the budget left for disagreement, so DE members can disagree much only if it is large or some nodes are poorly connected to the anchors.
The model enters the bound only through $\rho$ and $\varepsilon$, which depend on both its architecture and its training.
For example, an MLP ignores the graph, so its predictions need not be smooth and its energy budget can be large (\Cref{fig:line-graph-example}).
For graph transformers it can go either way: their $\rho$ depends on whether they \emph{learn} a smoothness bias.
For message passing, we expect $\rho$ to be small, as the next result suggests.

\begin{proposition}[Corollary 3.4.1 in \textcite{cai2020note}]
\label{prop:bound-on-rho}
For an $l$-layer GCN with ReLU activations and no biases, $\c{E}(\m{X}^{(l)}) = \c{O}((s \bar{\lambda})^l)$, where $\m{X}^{(l)}$ are the node representations after $l$ layers, $\c{E}$ is the energy for the augmented normalized Laplacian, $s$ bounds the squared spectral norms of the weight matrices, and $\bar{\lambda} = \max_i (1 - \lambda_i)^2 < 1$ over the nonzero eigenvalues $\lambda_i$ of this Laplacian.
\end{proposition}

Thus, the energy shrinks with depth when $s \bar{\lambda} < 1$, as in \Cref{fig:line-graph-example}.
Such results, also available for attention-based GNNs \parencite{Wu2023}, were developed to study oversmoothing, where the representations of different nodes within one network become indistinguishable after many layers \parencite{Oono2020,Rusch2023Oversmoothing}.\footnote{Beyond oversmoothing, prior work relates message passing to the variance of predictions, but only for infinitely wide networks (NTK) \parencite{Du2019gntk,Krishnagopal2023gntk,Hu2020infinitegnn,Niu2023gnnkernel} or at initialization \parencite{Li2023Virgo,Kelesis2025}, not for trained, finite networks.}
Our bound puts them to use in a much milder regime: it needs no oversmoothing, only a moderate budget $\rho$ at realistic depth, where smoothing is known to be beneficial \parencite{Keriven2022}.
This supports Assumption 1 for message-passing architectures, though it does not guarantee a small $\rho$, since $s$ depends on training.
However, recent results~\parencite{Keriven2026} seem to suggest $s$ remains generally small on message-passing models.

\paragraph{Discussion and limitations.}
Together, the theorem and the proposition above theoretically support the epistemic collapse: message passing tends to keep $\rho$ small, leaving little spare energy for disagreement.
Heterophily may leave more: a model that captures it must change its predictions between neighbors throughout the graph, which can raise $\rho$ well above $\c{E}_{\min}$.
This is consistent with the heterophilous \texttt{CHAM} having the largest NLL gain and epistemic-to-aleatoric ratio among regression tasks.
The bound does not account for training dynamics, so it cannot by itself explain the gap between a DE and a B-GCN of the same family on \texttt{PEMS} (\Cref{sec:motivation}).
Still, combined with our empirical findings, it offers an explanation: DE members each converge to different optima that implement nearly the same function (\Cref{sec:analysis:diversity}), whereas the B-GCN samples weights around an optimum, spreading its predictions within the limit set by the bound.

\section{Conclusion}\label{sec:conclusion}

We can now answer the question posed in \Cref{sec:motivation}.
The failure on \texttt{PEMS} is indeed a symptom of a systematic weakness of DEs of GNNs: epistemic collapse.
Empirically, the members of a DE barely disagree on every dataset we study.
The NLL gains of the DE are far smaller than in computer vision and come almost entirely from averaging point predictions.
Theoretically, we link this collapse to the smoothing bias of message passing (not \emph{over}smoothing!): it makes predictions similar for neighboring nodes.
For node-level tasks, our bound shows that such smoothness, together with training fit, limits how much the members can disagree.
Together, our empirical and theoretical results answer the question in our title: \emph{Do deep ensembles actually capture uncertainty in graph neural networks?}
Largely, they do not, at least not beyond what a single member already captures.

In practice, standard DEs of GNNs multiply training and inference costs, which is hard to justify given that they mostly average out optimization noise.
Since the best of several seeds can come close to the ensemble, selecting it on validation might retain most of this benefit at single-model inference cost.
Better uncertainty on graphs is nonetheless attainable: on \texttt{PEMS}, a Bayesian GCN beats the Trivial baseline, so the smoothing bias limits, but does not altogether preclude, useful epistemic uncertainty in GNNs.
We studied standard DEs of message-passing GNNs because their appeal lies in simplicity: they only require training the same model several times.
Graph-specific diversity techniques or architectures beyond message passing may fare better at the cost of this simplicity, and we hope our analysis guides the search for methods that deliver better uncertainty on graphs.

\newpage

\clearpage
\newpage

\subsection*{Acknowledgements}
Pedro C. Vieira and Pedro Ribeiro are funded by national funds through FCT – Fundação para a Ciência e a Tecnologia, I.P., under the PhD scholarship \href{https://doi.org/10.54499/2025.02316.BD}{2025.02316.BD} and under the grant UID/50014/2025.

\subsection*{AI use statement}

In this work, we used generative AI tools to review and give feedback on the theoretical results and their proofs, and for brainstorming.
We have not used generative AI tools for the other tasks requiring disclosure, such as generating data, implementing methods, or running experiments.
Additionally, we used generative AI tools to edit the text, to aid in creating figures, to aid the literature review, and for occasional code verification.
We have reviewed all AI-assisted work by at least two authors, and checked cited results against the original papers.
We take responsibility for the final content of this work, including text, claims or artifacts produced with the aid of generative AI.

\subsection*{Reproducibility statement}

Complete hyperparameter configurations, baseline details, and experimental setups are provided in \Cref{sec:appx-extra-details-exp}.
We also cite relevant works that describe the setups we adopted in more detail.
\Cref{sec:appx-theory} states all assumptions and gives detailed proofs of our theoretical results.
We provide the code used to run the experiments as supplemental material.
It includes instructions on how to run the experiments and anonymous links to intermediary results for inspection.
All code uses standard rootless containerized environments to ensure ease of use and reproducibility.

\clearpage
\newpage

\printbibliography

\clearpage
\newpage

\appendix
\crefalias{section}{appendix}

\section{Extra Details on the Experiments Conducted}\label{sec:appx-extra-details-exp}

\textit{Like in the main text each experiment is repeated ten times to obtain the required statistics.
Furthermore, for all plots in these sections, the marker indicates the mean improvement. The thick whisker denotes $\pm \sigma$, and the thin semi-transparent lines represent the minimum and maximum values.}

The selected hyperparameters used throughout the experiments in the main text are displayed in~\Cref{tab:hyperparameters}.
For the GAT models, the number of attention heads used was four.
For Tolokers2, Artnetviews and Chameleon, besides the GNN we used a multi-layer perceptron (MLP) module to pre-process the input.
We also apply an MLP in between the GNN layers and after the output of the final GNN layer.
The MLP used has depth two.
The models for QM9 and MolHIV also include a pooling mechanism to create the graph level representations.
This mechanism concatenates the representations obtained from the basic pooling mechanisms: mean, maximum and standard deviation.
This representation is then used as input for a two layer MLP with 256 hidden dimension that will produce the final graph level representation.
Additionally, MolHIV uses a step LR scheduler with step size 10 and $\gamma = 0.707$, and the standard OGB scaffold split.

\begin{table}[ht]
\centering
\caption{Hyperparameter configurations used the used datasets.}
\label{tab:hyperparameters}
\resizebox{\textwidth}{!}{
\begin{tabular}{llcccccc}
\toprule
\textbf{Dataset} & \textbf{Model} & \textbf{Hidden Dim.} & \textbf{Layers} & \textbf{Learning Rate} & \textbf{Weight Decay} & \textbf{Epochs} & \textbf{Dropout} \\
\midrule
Tolokers2 & GATv2 & 256 & 3 & 0.0003 & 0.0000      & 1000 & 0.1 \\
Citeseer & GCN   & 20  & 2 & 0.0100 & 0.0005 & 200  & 0.0\\
MolHIV   & GIN   & 64  & 2  & 0.0010  & 6e-6   & 60  & 0.525   \\
Cora (\Cref{sec:appx-extra-plots}) & GCN   & 20  & 2 & 0.0100 & 0.0005 & 500 & 0.0  \\
Artnetviews & GCN   & 512 & 3 & 0.0003 & 0.0000     & 1000 & 0.1 \\
Chameleon & GATv2 & 512 & 3 & 0.0003 & 0.0000      & 250 & 0.3  \\
QM9 & GCN   & 128 & 4 & 0.0010 & 0.0000      & 500  & 0.1 \\
PEMS & GCN   & 100 & 3 & 0.0100 & 0.0005      & 500 & 0.0 \\
\bottomrule
\end{tabular}
}
\end{table}

The parameters for~\Cref{sec:motivation}, GCN and GCN DE follow~\Cref{tab:hyperparameters}.
The Bayesian GCN was trained using SVI~\parencite{Blundell2015,Graves2011,Louizos2017,Louizos2016,Wen2018} with a Normal surrogate distribution.
The prior was initialized with standard Glorot initialization~\parencite{glorot2010-init} scaled by 177.7.
The hidden dimension was reduced to 50 to speed convergence.
The Matérn GP followed~\textcite{borovitskiy2025}\footnote{All experiments were performed using a NVIDIA RTX 5090 or equivalent.}.

\section{Dataset Shift}\label{sec:appx-data-shift}

Besides general uncertainty quantification in in-distribution (ID) settings, uncertainty quantification is popular in out-of-distribution (OOD) settings~\parencite{Ovadia2019}.
Like in ID settings it is important that a model can signal if it is uncertain about the prediction it makes.
However, in OOD settings, the model is acting on data that is likely no longer supported by the training distribution.
Thus evaluating uncertainty quantification in OOD scenarios is also very important since we expect cases where uncertainty increases to appear more often than in ID.
To extend our findings to OOD we selected two datasets, one for classification (Tolokers2) and one for regression (Artnetviews), and split them into OOD and ID.
Since all results held across datasets for ID, we selected the two largest node datasets as representatives of the others.

To evaluate model robustness under distribution shift, we adopt the density-based split proposed by~\textcite{Bazhenov2023-graphshift}.
This method partitions data based on local graph topology rather than feature distribution.
However, it still introduces meaningful partitions in graph datasets, often being able to replicate splits based on features~\parencite{Bazhenov2023-graphshift}.

Importantly, the number of training, validation, and test examples is kept identical to the original fully ID split.
We first rank all nodes by their local clustering coefficient in descending order. The first $N_\text{test}$ examples, where $N_\text{test}$ is the number of test examples in ID, are assigned for test, followed by $N_\text{val}$ nodes for validation and $N_\text{train}$ nodes for training.
Consequently, test nodes correspond to those with the highest local clustering coefficients, while training nodes have the lowest.
This split based on the connectedness of a node's neighbouhood assigns denser neibhourhoods, the ones that can capture richer information, almost exclusively to the test set.

\begin{figure}[h]
    \centering
    \begin{subfigure}[b]{0.48\textwidth}
        \centering
        \includegraphics[width=\linewidth]{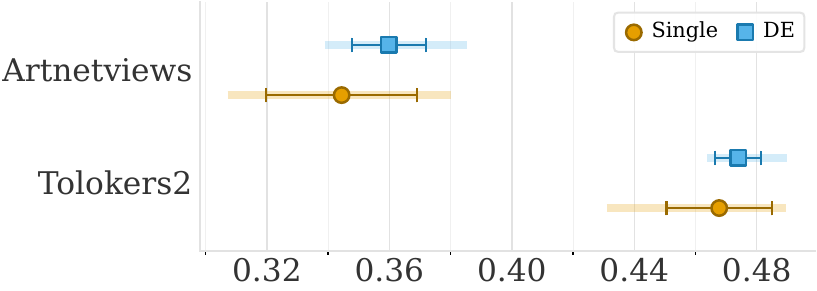}
        \caption{Point Estimation}
        \label{fig:ood:pe:appendix}
    \end{subfigure}
    \hfill
    \begin{subfigure}[b]{0.48\textwidth}
        \centering
        \includegraphics[width=\linewidth]{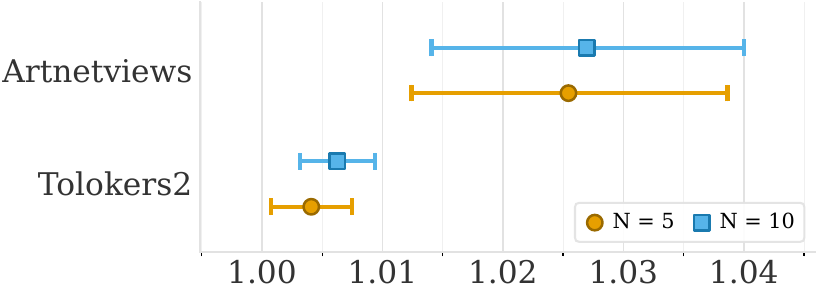}
        \caption{Likelihood Ratio}
        \label{fig:ood:ratio:appendix}
    \end{subfigure}
    
    \vspace{0.5em}
    \caption{Comparison of point estimation performance, $R^2$, for Artnetviews and average precision for Tolokers2, between a DE (5 models) and a single model. For both metrics higher score is better (\subref{fig:ood:pe:appendix}). Likelihood ratio \( \exp\del{\text{NLL}_{\text{GNN}} - \text{NLL}_{\text{DE}}} \) and 95\% CI assuming normally distributed errors, quantifying the relative predictive likelihood improvement of the DE (5 models) over a single model in structural OOD shift (\subref{fig:ood:ratio:appendix}).}
    \label{fig:ood:appendix}
\end{figure}

As expected, the performance in point estimation drops between ID and OOD.
Specifically, for a five model ensemble for Artnetviews we register a drop in $R^2$ from $0.599 \pm 0.003$ to $0.360 \pm 0.012$ and for Tolokers2 a drop in average precision from $0.617 \pm 0.003$ to $0.474 \pm 0.007$\footnote{We use $R^2$ and average precision to enable easy comparison with the original work by~\textcite{bazhenov2025graphland} if necessary.}.
However, like the ID case, deep ensembles provide little gains over a single model for uncertainty quantification.
~\Cref{fig:ood:appendix} shows that the gains of an ensemble for Tolokers2 are $<1\%$ and for Artnetviews $<3\%$---an even worse performance than ID for Artnetviews.

\section{Model Soup}\label{sec:appx-model-soup}

\begin{figure}[t]
    \centering
    \begin{subfigure}[b]{0.48\textwidth}
        \centering
        \includegraphics[width=\linewidth]{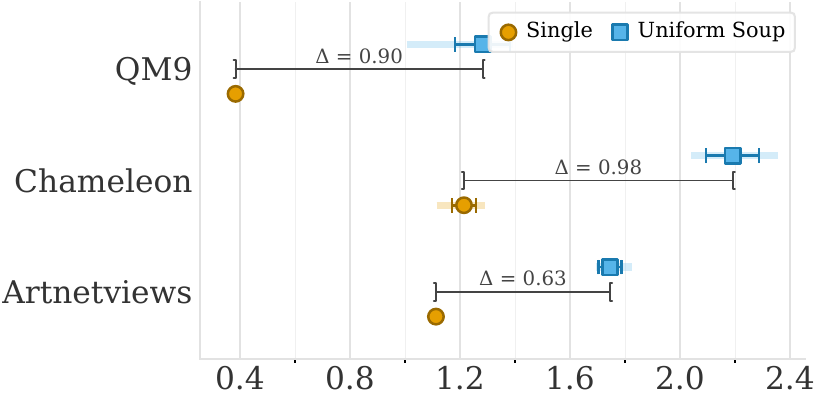}
        \caption{Model Soup --- Point Estimation (RMSE)}
                \label{fig:convexity:appendix:pe:reg}
    \end{subfigure}
    \hfill
    \begin{subfigure}[b]{0.48\textwidth}
        \centering
        \includegraphics[width=\linewidth]{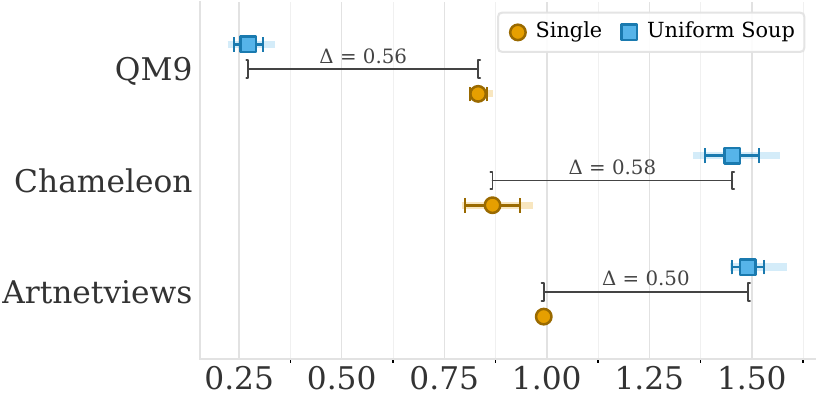}
        \caption{Model Soup --- NLL}
        \label{fig:convexity:appendix:nll:reg}
    \end{subfigure}
    
    \vspace{0.5em}
    \caption{Weight-space convexity analysis.
    (\subref{fig:convexity:appendix:pe:reg}) compares the point prediction performance and (\subref{fig:convexity:appendix:nll:reg}) compares the NLL of a single model against a model soup formed by averaging the weights of the ensemble members.
    The degraded performance of the model soup confirms that the models reside in different regions of a highly non-convex weight space. Markers represent the mean, thin bars represent $\pm\sigma$ and transparent bars represent min--max.}
    \label{fig:convexity:appendix:reg}
\end{figure}

\begin{figure}[h]
    \centering
    \begin{subfigure}[b]{0.48\textwidth}
        \centering
        \includegraphics[width=\linewidth]{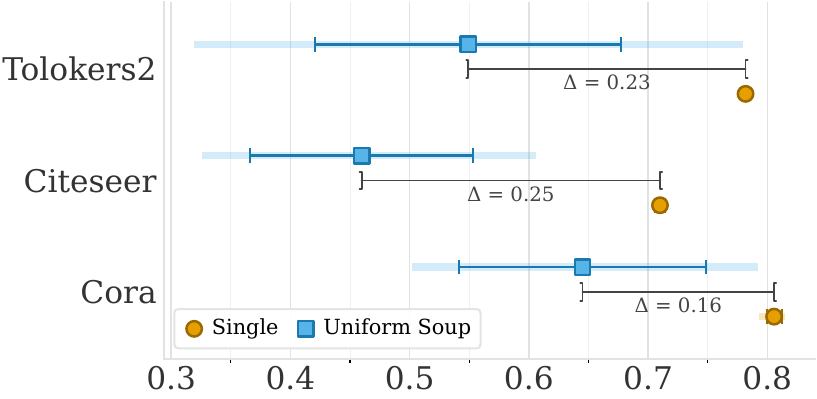}
        \caption{Model Soup -- Point Estimation}
                \label{fig:convexity:pe:appendix:class}
    \end{subfigure}
    \hfill
    \begin{subfigure}[b]{0.48\textwidth}
        \centering
        \includegraphics[width=\linewidth]{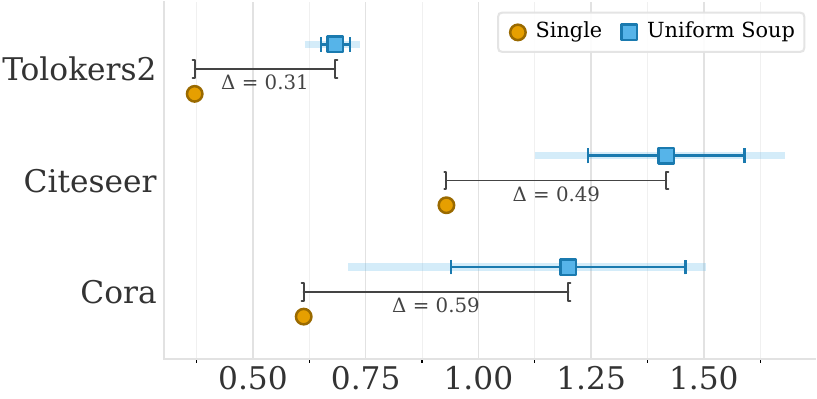}
        \caption{Model Soup -- NLL}
        \label{fig:convexity:nll:appendix:class}
    \end{subfigure}
    
    \vspace{0.5em}
    \caption{Weight-space convexity analysis.
    \Cref{fig:convexity:pe:appendix:class} compares the point prediction performance and \Cref{fig:convexity:nll:appendix:class} the NLL of a single model against a "model soup" formed by averaging the weights of the ensemble members. Lower values are better for NLL and higher are better for point prediction.
    The degraded performance of the model soup confirms that the models reside in different regions of a highly non-convex parameter space.}
    \label{fig:convexity:appendix:class}
\end{figure}

As noted in \Cref{sec:analysis:diversity}, we construct a "model soup" \parencite{wortsman2022} by averaging the weights of the individual ensemble members.
If the models resided within a simple convex basin in the weight space, averaging their weights would act as a variance reduction mechanism over a noisy approximation of the global optimum, yielding a point closer to the true global optimum and thus improving performance.
However, as shown in \Cref{fig:convexity:appendix:reg}, \Cref{fig:convexity:appendix:class}, the model soup generally underperforms a single model.

For regression, specifically for point prediction, its performance degrades to essentially match a Trivial baseline.
In the only case where the aggregate NLL metric technically improves (\texttt{QM9-5\%}), the soup remains practically uninformative for uncertainty quantification, as it outputs near-identical uncertainties for all predictions.
This degraded behavior confirms that the weight space itself remains highly non-convex---a natural consequence once one considers the permutation symmetries inherent in deep neural networks \cite{entezari2022,ainsworth2023}.

Like the results for regression, model soup deteriorates performance for both point prediction and uncertainty\footnote{For classification, the model soup uses Cora (\Cref{sec:appx-extra-plots}) in place of \texttt{MHIV}.}.
Despite the inherent limitations of accuracy, for example, the insensitivity to confidence and the threshold rigidity even the best model soup cannot reach the worst single model.
Furthermore, the gaps in NLL are more pronounced than those observed for accuracy, indicating that model soup performs particularly poorly in terms of uncertainty estimation.
In conclusion, even if model soup consistently matched the performance of a single model, this would still be insufficient evidence for weight-space convexity. 
To support that hypothesis, model soup would need to clearly outperform the single model and at least match ensemble performance~\parencite{wortsman2022}.
The closer the gap from the model soup to the ensemble performance the stronger the weight-space convexity hypothesis holds (assuming standard optimization procedures without perfect convergence to the global optimum).
In our results, model soup fails to match even the single-model baseline.
Thus we argue against weight-space convexity.

Therefore, we are left with a compelling hypothesis: the weight space is non-convex, yet epistemic collapse strongly suggests that GNNs experience a form of convexity in the \emph{functional} space.
That is, independently trained models find different sets of weights that ultimately implement the same (or similar) predictive function.

\section{Proof of Theorem in \texorpdfstring{\Cref{sec:theory}}{Section \ref{sec:theory}}}
\label{sec:appx-theory}

In this section we prove the main theorem from~\Cref{sec:theory} and extend it to models that achieve perfect training performance and for graphs with multiple connected components.

\subsection{Setup and assumptions}

Let $G=(V,E)$ be a finite, connected, undirected, unweighted graph with $|V|=n$ and combinatorial Laplacian $\m{L} = \m{D} - \m{A}$.
The results derived extend to the other Laplacian, for example, symmetric normalized.
We chose the combinatorial for convenience.
Let $\varnothing \neq S\subsetneq V$ be the labeled nodes, with fixed labels $\v{y}_S$, and let $U=V\setminus S$ be unlabeled (test) nodes.
For $z\in\R^n$, the Dirichlet energy is\footnote{Contrarily to the main text,~\Cref{sec:theory}, for visual clarity, we will distinguish between minimum Dirichlet Energy for non-perfect training error, $\mathcal{E}_{\text{min}}^{\epsilon}$, and perfect training error $\mathcal{E}_{\text{min}}$. The main text refers to the former.}
\[
\mathcal{E}(\v{z})=\v{z}^\top \m{L} \v{z}
=\sum_{\{i,j\}\in E}(\v{z}_i-\v{z}_j)^2
\]

Let $\v{f}_w$ be the prediction vector obtained from a randomly initialized (e.g., Glorot) trained model.
We assume a distribution over trained weights $w \sim \mu$, $\E_w[\|\v{f}_w\|_2^2]<\infty$, and define
\[
\bar{\v{f}}=\E_w[\v{f}_w],
\qquad
\v{e}=\v{f}_w-\bar{\v{f}},
\qquad
V_{\text{epi}}
=\sum_{v\in V}\Var_w((\v{f}_w)_v)
=\E_w[\|\v{e}\|_2^2]
\]

\noindent\textbf{Assumption 1 (Expected prediction energy).}
For some finite $\rho$,
\[
\E_w[\mathcal{E}(\v{f}_w)]\leq\rho
\]

\noindent\textbf{Assumption 2 (Bounded squared error).}
For every $s\in S$, there are $\varepsilon_s \geq 0$ such that
\[
\E_w[(\v{f}_w(s) - \v{y}_s)^2] \leq \varepsilon^2_s
\]
Because $\v{y}_s$ is fixed, we note that the equation above can be interpreted as bounding the prediction variance by $\varepsilon^2_s$ and expected deviation from the prediction to the label by $\varepsilon_s$. 

\paragraph{Why prediction energy alone is insufficient.}
Take any fixed $\v{h}$ with $\v{h}_S=\v{y}_S$ and let $\v{f}_w = \v{h}+Z\mathbf{1}$, where $Z$ is a scalar random variable with finite but arbitrarily large variance.
Since $\m{L}\mathbf 1=0$, if we assume $\E[Z] = \mu_Z = 0$, taking $\bar{\v{f}} = \E_w[\v{f}_w] = h+\mu_Z\v{1}$, yields

\[
\v{e}
&= \v{f}_w - \bar{\v{f}}\\
&= \v{h}+Z\v{1} - (\v{h}+\mu_Z\v{1})\\
&= (Z-\mu_Z)\v{1}
\]

leading to $V_{\text{epi}} = \E_w[\|\v{e}\|_2^2] = n \Var(Z)$ while $\mathcal{E}(\v{f}_w) = \mathcal{E}(\v{h})$ almost surely.
Thus an energy bound cannot, by itself, bound epistemic variance.
Assumption 2 controls this example because $\Var_w((\v{f}_w)_s)=\Var(Z)$ at every labeled node.

\subsection{Spectral and minimum-energy lemmas}

For $A,B\subseteq V$, let $\m{L}_{AB}=(\m{L}_{ij})_{i\in A,j\in B}$; in particular, $\m{L}_{UU}$ is a principal submatrix of the \emph{full} Laplacian.
Its diagonal entry at $u\in U$ is the degree of $u$ in $G$, including edges from $u$ to $S$.
Equivalently, if $\m{L}_U^{\text{ind}}$ is the Laplacian of the subgraph induced by $U$ and $a_{us}=\mathbf{1}_{\{\{u,s\}\in E\}}$, then

\[
\m{L}_{UU}
=
\m{L}_U^{\text{ind}}
+\operatorname{diag}\left(\sum_{s\in S}a_{us}\right)_{u\in U}
\]

We refer to this principal submatrix as the grounded Laplacian.

Let $\m{P}_S$ be the diagonal projector onto the coordinates in $S$, so $\v{x}^\top \m{P}_S \v{x}=\|\v{x}_S\|_2^2$.
For $\alpha>0$, define

\[
\m{M}_\alpha=\m{L}+\alpha \m{P}_S,
\qquad
\lambda_\alpha=\lambda_{\min}(\m{M}_\alpha)
\]

\begin{lemma}[Anchored and grounded spectral inequalities]
\label{lem:anchored-spectral:appx}
Both $\m{M}_\alpha$ and $\m{L}_{UU}$ are positive definite.
Consequently, for every $\v{x}\in\R^n$,
\[
\|\v{x}\|_2^2
\leq
\frac{\v{x}^\top \m{L} \v{x} + \alpha\|\v{x}_S\|_2^2}{\lambda_\alpha}.
\]
If $\v{x}_S=0$, then also
\[
\|\v{x}\|_2^2
\leq
\frac{\v{x}^\top \m{L} \v{x}}{\lambda_{\min}(\m{L}_{UU})}
\]
\end{lemma}

\begin{proof}
\textit{1st Inequality}:
Both terms in $\v{x}^\top \m{M}_\alpha \v{x} = \v{x}^\top \m{L} \v{x} + \alpha\|\v{x}_S\|_2^2$ are nonnegative.
If their sum is zero, connectedness implies $\v{x} \in \ker \m{L}=\operatorname{span}\{\mathbf{1}\}$; the nonempty set $S$ then forces $\v{x}=0$.
Thus $\m{M}_\alpha \succ 0$, and its Rayleigh quotient gives the first Poincaré inequality~\parencite{Fan1997Spectral}.

\textit{2nd Inequality}:
If $\v{z} \in \R^{|U|}$, extend it to $\v{x} \in \R^n$ by setting
$\v{x}_S=0$ and $\v{x}_U=\v{z}$.
Then $\v{z}^\top \m{L}_{UU} \v{z} = \v{x}^\top \m{L} \v{x}$.
This is zero only if $\v{x}$ is constant, hence, because $\v{x}_S=0$, only if $\v{z}=0$. Therefore $\m{L}_{UU}\succ0$.
For any $\v{x}$ with $\v{x}_S=0$, we have $\v{x}^\top \m{L} \v{x} = \v{x}_U^\top \m{L}_{UU} \v{x}_U$ and $\|\v{x}\|_2=\|\v{x}_U\|_2$; the second Poincaré inequality~\parencite{Fan1997Spectral} follows from the Rayleigh quotient of $\m{L}_{UU}$.
\end{proof}

For fixed labeled values $\v{b}\in\R^{|S|}$, define
\[
\mathcal{E}_{\min}(\v{b})
=
\min\{\v{z}^\top \m{L} \v{z} : \v{z}_S = \v{b}\}
\]
For the interpolation-error intervals, define
\[
\mathcal{E}_{\min}^{\epsilon}(\v{y}_S)
=
\min\left\{\v{z}^\top \m{L} \v{z}:
|\v{z}_s-\v{y}_s| \leq \varepsilon_s \ \text{for every } s\in S\right\}.
\]

\begin{lemma}[Minimum energy with prescribed labeled values]
\label{lem:minimum-energy:appx}
Let

\[
\m{R}=\m{L}_{SS} - \m{L}_{SU} \m{L}_{UU}^{-1} \m{L}_{US}
\]

Then

\[
\mathcal{E}_{\min}(\v{b}) = \v{b}^\top \m{R} \v{b},
\qquad
\mathcal{E}_{\min}^{\epsilon}(\v{y}_S)
=
\min_{\substack{\v{b} \in \R^{|S|} \\
                 |\v{b}_s-\v{y}_s| \leq \varepsilon_s\ \forall s\in S}}
\v{b}^\top \m{R} \v{b}
\]

For fixed $\v{b}$, the minimum is attained by the harmonic extension

\[
\v{z}_S=\v{b},\qquad \v{z}_U=-\m{L}_{UU}^{-1}\m{L}_{US}\v{b}
\]
\end{lemma}

\begin{proof}
By \Cref{lem:anchored-spectral:appx}, $\m{L}_{UU}$ is invertible.
Completing the square, with $\v{z}_S=\v{b}$, gives

\[
\v{z}^\top \m{L} \v{z}
=
\bigl(\v{z}_U + \m{L}_{UU}^{-1} \m{L}_{US} \v{b}\bigr)^\top
 \m{L}_{UU}
\bigl(\v{z}_U + \m{L}_{UU}^{-1} \m{L}_{US}\v{b}\bigr)
+\v{b}^\top \m{R} \v{b}
\]

Minimizing first over $\v{z}_U$ proves the formula for $\mathcal{E}_{\min}(\v{b})$.
Minimizing subsequently over the compact box of admissible $\v{b}$ proves the second formula and attainment of the minimum.
\end{proof}

\begin{lemma}[Centered-energy identity]
\label{lem:centered-energy:appx}
The centered predictions satisfy

\[
\E_w[\v{e}^\top \m{L} \v{e}]
=
\E_w[\v{f}_w^\top \m{L} \v{f}_w]
-\bar{\v{f}}^\top \m{L} \bar{\v{f}}.
\]

Under Assumptions 1 and 2, it follows that

\[
\E_w[\v{e}^\top \m{L} \v{e}]
\leq
\rho-\mathcal{E}_{\min}^{\epsilon}(\v{y}_S)
\]

\end{lemma}

\begin{proof}
Expand $\v{f}_w=\bar{\v{f}}+\v{e}$ and use $\E_w[\v{e}]=0$ to obtain the identity.
Assumption 2 places $\bar{\v{f}}_S$ in the prescribed bounded squared-error interval (e.g., by Jensen's inequality $|\bar{\v{f}}_s-\v{y}_s|^2 \le \E_w(\v{f}_w(s)-\v{y}_s)^2\le\varepsilon_s^2$).
Hence $\bar{\v{f}}^\top \m{L}\bar{\v{f}} \geq \mathcal{E}_{\min}^{\epsilon}(\v{y}_S)$, while Assumption 1 bounds the first term in the identity by $\rho$.
\end{proof}

\subsection{Variance bounds on a connected graph}

\begin{theorem}[Soft-Pinning Epistemic Variance Bound]
\label{thm:soft-bound-epistemic-variance:appx}

Under Assumptions 1 and 2,

\[
\label{eq:soft-bound-epistemic-variance:appx}
\boxed{
V_{\text{epi}}
\leq
\inf_{\alpha>0}
\frac{
\rho-\mathcal{E}_{\min}^{\epsilon}(\v{y}_S)
+\alpha\sum_{s\in S}\varepsilon_s^2
}{
\lambda_\alpha
}}
\]

\end{theorem}

\begin{proof}
Apply the first inequality of \Cref{lem:anchored-spectral:appx} to $\v{e}$ and take expectations:

\[
V_{\text{epi}}
\leq
\frac{
\E_w[\v{e}^\top \m{L} \v{e}]
+ \alpha\E_w[\|\v{e}_S\|_2^2]
}{\lambda_\alpha}.
\]

By \Cref{lem:centered-energy:appx}, the first term in the numerator is at most $\rho-\mathcal{E}_{\min}^{\epsilon}(\v{y}_S)$.
By Assumption 2,
\[
\E_w[\|\v{e}_S\|_2^2]
=
\sum_{s\in S}\Var_w((\v{f}_w)_s)
\leq\sum_{s\in S}\varepsilon_s^2
\]

The resulting bound holds for every $\alpha>0$; take the infimum.
\end{proof}

\begin{corollary}[Hard-Pinning Variance Bound]
\label{corollary:hard-pinning:appx}

Suppose $\v{e}_S=0$ almost surely, so all models have the same labeled predictions $\v{b}_S=\bar{\v{f}}_S$, not necessarily equal to $\v{y}_S$.
Under Assumption 1,

\[
\label{eq:hard-bound-epistemic-variance:appx}
\boxed{
V_{\text{epi}}
\leq
\frac{\rho-\mathcal{E}_{\min}(\v{b}_S)}
{\lambda_{\min}(\m{L}_{UU})}}
\]

In particular, exact interpolation corresponds to $\v{b}_S=\v{y}_S$.
\end{corollary}

\begin{proof}

Since $\v{e}_S=0$ almost surely, the grounded inequality in \Cref{lem:anchored-spectral:appx} gives

\[
V_{\text{epi}}
\leq
\frac{\E_w[\v{e}^\top \m{L}\v{e}]}{\lambda_{\min}(\m{L}_{UU})}
\]

The centered-energy identity holds independently of Assumption 2.
As $\bar{\v{f}}_S=\v{b}_S$, we have $\bar{\v{f}}^\top \m{L} \bar{\v{f}} \geq\mathcal{E}_{\min}(\v{b}_S)$, and Assumption 1 therefore gives $\E_w[\v{e}^\top \m{L}\v{e}]\leq\rho-\mathcal{E}_{\min}(\v{b}_S)$.
\end{proof}

\subsection{Extension to disconnected graphs}

Now let $G$ have connected components $C_1,\ldots,C_K$, and set $S_k=S\cap C_k$ and $U_k=U\cap C_k$.
A component with no labeled node has an uncontrolled constant mode, so an additional condition is necessary.

\noindent\textbf{Assumption 3 (A label in every component).} $S_k\neq\varnothing$ for every $k$.

Let $\m{L}^{(k)}$ be the Laplacian of the component $C_k$, and let $P_{S_k}$ be its labeled-coordinate projector. Define

\[
\m{M}_{\alpha,k}=\m{L}^{(k)}+\alpha \m{P}_{S_k},
\qquad
\lambda_{\alpha,k}=\lambda_{\min}(\m{M}_{\alpha,k})
\]

and define $\mathcal{E}_{\min,k}^{\epsilon}((\v{y}_S)_k)$ using the edges and labeled nodes of $C_k$. Choose any constants $\rho_k$ satisfying

\[
\E_w\bigl[
(\v{f}_w)_{C_k}^{\top}\m{L}^{(k)}(\v{f}_w)_{C_k}
\bigr]\leq\rho_k.
\]

For example, each $\rho_k$ may be taken equal to the expectation on the left; their sum is then at most $\rho$ under Assumption 1.

\begin{corollary}[Disconnected-Graph Soft-Pinning Variance Bound]
\label{corollary:general-soft-pinning:appx}
Under Assumptions 1--3,

\[
\label{eq:general-soft-bound-epistemic-variance:appx}
\boxed{
V_{\text{epi}}
\leq
\sum_{k=1}^K
\inf_{\alpha_k>0}
\frac{
\rho_k-\mathcal{E}_{\min,k}^{\epsilon}((\v{y}_S)_k)
+\alpha_k\sum_{s\in S_k}\varepsilon_s^2
}{
\lambda_{\alpha_k,k}
}}
\]

The formula also covers components with $U_k=\varnothing$.
\end{corollary}

\begin{proof}
Each $C_k$ is connected and contains a labeled node.
Thus the positive-definiteness, centered-energy, and soft-pinning arguments apply to that component, yielding the $k$-th summand as a bound on $\E_w[\|\v{e}_{C_k}\|_2^2]$.
Sum these component-wise bounds and use $V_{\text{epi}}=\sum_k\E_w[\|\v{e}_{C_k}\|_2^2]$.
The penalty strengths may be optimized separately because the component-wise inequalities are independent.
\end{proof}

In particular, a component with few labeled nodes may have fewer "anchors", meaning lower minimum energy and a looser bound than components with a large amount of train nodes.
Similarly, a component with low expansion coefficient or when unlabeled nodes are poorly connected to the training nodes, for instance when they are far from them or reachable only through bottlenecks, lead to a looser bound due to smaller $\lambda_{\alpha,k}$.

\subsection{Interpretation of the expected-energy assumption}

The quantity $\rho$ is an upper bound on Dirichlet energy, or equivalently, on the \emph{expected squared differences between predictions at adjacent nodes}:

\[
\E_w[\mathcal{E}(\v{f}_w)]
=
\sum_{\{i,j\}\in E}
\E_w\left[
\bigl((\v{f}_w)_i-(\v{f}_w)_j\bigr)^2
\right]
\]

Representation-collapse results such as those discussed by \textcite{Ma2022Homophily} can motivate expecting small prediction differences on some same-class edges, but feature-space closeness or a high-probability collapse statement does not alone provide an upperbound on $\rho$.
Such a deduction would also require control of the readout from representations to predictions and of the squared differences on the remaining edges.
We select an upperbound in~\Cref{prop:bound-on-rho}.

Good heterophily can support accurate message passing without graph
homophily \parencite{Ma2022Homophily,Luan2024Heterophilic,Luan2026Heterophilic}.
By \Cref{lem:centered-energy:appx}, however, epistemic variance is controlled not by $\rho$ itself but by the fluctuation energy
\[
\E_w[\v{e}^\top \m{L} \v{e}]
=
\E_w[\mathcal{E}(\v{f}_w)]-\bar{\v{f}}^\top \m{L}\bar{\v{f}},
\]
which our bounds estimate from above by the spare energy $\rho-\mathcal{E}_{\min}^{\epsilon}(\v{y}_S)$.
If label changes across edges are concentrated among unlabeled nodes (e.g., heterophily between test nodes but not between training nodes), a model that follows this structure may require a larger $\rho$, while $\mathcal{E}_{\min}^{\epsilon}(\v{y}_S)$, which depends only on the training labels, is unchanged.
The spare energy, and hence the bound, then grows.

Two caveats limit what this implies.
First, a looser upper bound permits, but does not force, larger epistemic variance.
Second, the bound lower-estimates $\bar{\v{f}}^\top \m{L}\bar{\v{f}}$ using the training labels only.
If the ensemble mean is also accurate on unlabeled nodes, the same argument yields the spare energy $\rho-\mathcal{E}_{\min}^{\epsilon}(\v{y}_V)$, and heterophily may raise both terms.
Room for disagreement therefore arises only insofar as individual members are rougher than their mean, something that should be observed in practice for heterophilic graphs.
That is, disagreement arises when ensemble members fit high-energy structure inconsistently rather than all capturing it.

Consequently, our framework does not predict that heterophily by itself
causes ensemble diversity.
It predicts that the constraint behind epistemic collapse binds most tightly when the task requires low-energy predictions, as on strongly homophilous graphs, and can be relaxed when the task requires high-energy predictions that message passing represents inconsistently across initializations.
We regard the latter as a plausible mechanism for the weaker collapse observed on \texttt{CHAM}, not as a direct consequence of the theorem.

\section{Other Plots and Results}\label{sec:appx-extra-plots}

\paragraph{More Datasets: Cora.} In addition to the datasets presented in the main text, we also tested the classic Cora (\texttt{CORA}) dataset~\parencite{Yang2016CoraCite}.
We followed the same reasoning as CiteSeer to pick the architecture.
Thus, we picked GCN for the better training stability it offers compared to GAT, despite the slightly worse point estimation performance.
Epistemic/aleatoric ratio is comparable to the ones shown in~\Cref{fig:uq-decomposition:class}; the same applies for Jensen's gap.
The likelihood ratios are slightly above the ones from~\Cref{fig:main_results_plot:classification} with mean around 1\% and confidence interval between 0.05\% and 2\%.

\begin{table}[h]
\centering
\caption{Absolute performance metrics for a DE of five models on Cora. Values represent NLL, ECE, and PEM (accuracy for classification), showing the mean and standard deviation over 10 independent runs. The ECE of the trivial baseline is not taken into account when highlighting the best score. Trivial baseline has low ECE by construction but is uninformative: it predicts the same regardless of input.}
\begin{adjustbox}{max width=\linewidth}
\begin{tabular}{cl ccc}
\toprule
\textbf{Dataset} & \textbf{Metric} & \textbf{Trivial} & \textbf{GNN} & \textbf{DE} \\
\midrule
\multirow{3}{*}{\texttt{CORA}} & NLL & $1.946 \pm 0.000$ & $0.612 \pm 0.009$ & $\boldsymbol{0.600 \pm 0.008}$ \\
& ECE & $0.116 \pm 0.000$ & $0.123 \pm 0.001$ & $\boldsymbol{0.122 \pm 0.002}$ \\
& PEM & $0.130 \pm 0.000$ & $0.806 \pm 0.006$ & $\boldsymbol{0.808 \pm 0.003}$ \\
\bottomrule
\end{tabular}
\end{adjustbox}
\label{tab:cora_appendix}
\end{table}

\begin{figure}[ht]
    \centering
    \begin{subfigure}{0.49\textwidth}
        \centering
        \includegraphics[width=\linewidth]{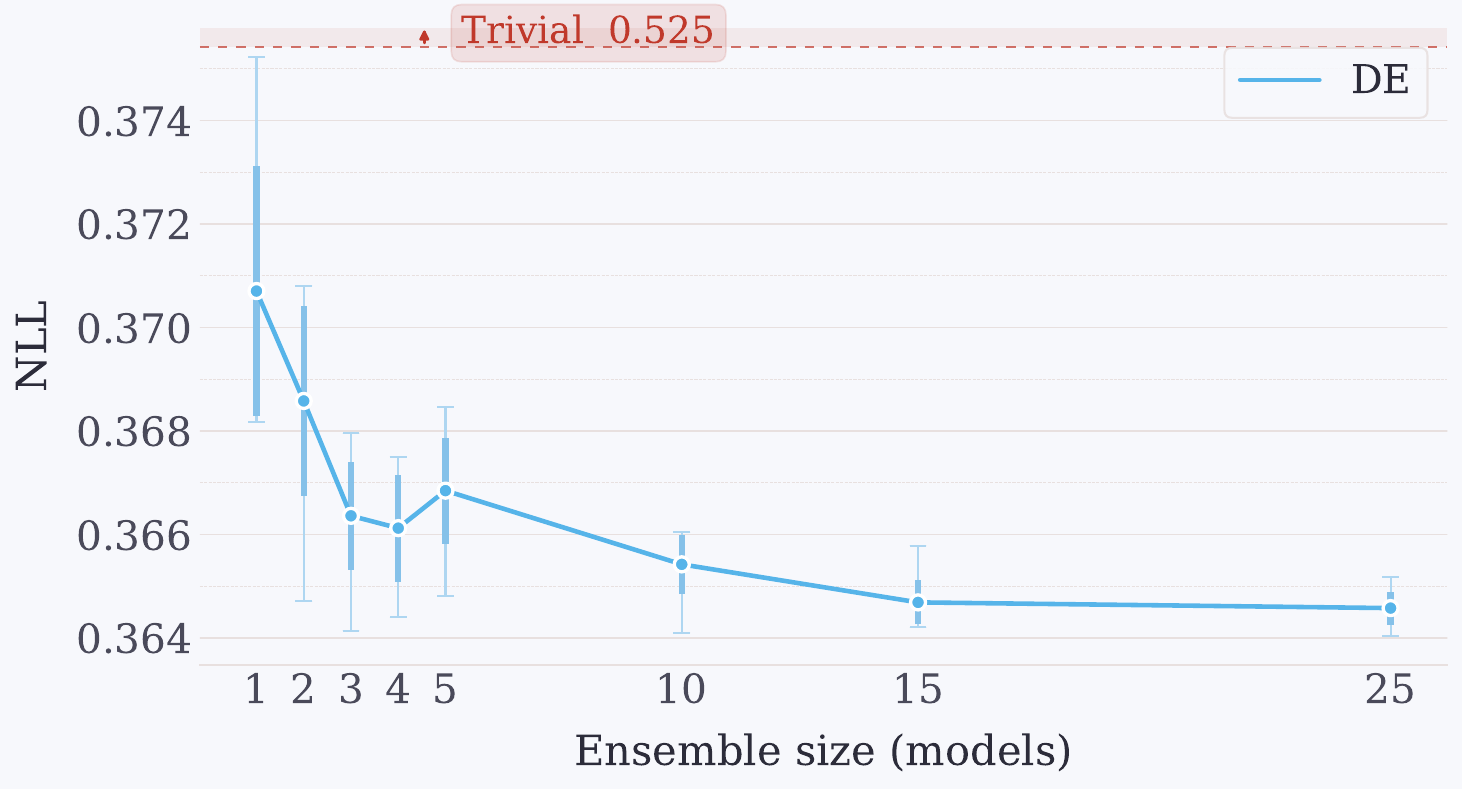}
        \caption{Tolokers2 NLL}
        \label{fig:tolokers2_nll}
    \end{subfigure}%
    \hfill
    \begin{subfigure}{0.49\textwidth}
        \centering
        \includegraphics[width=\linewidth]{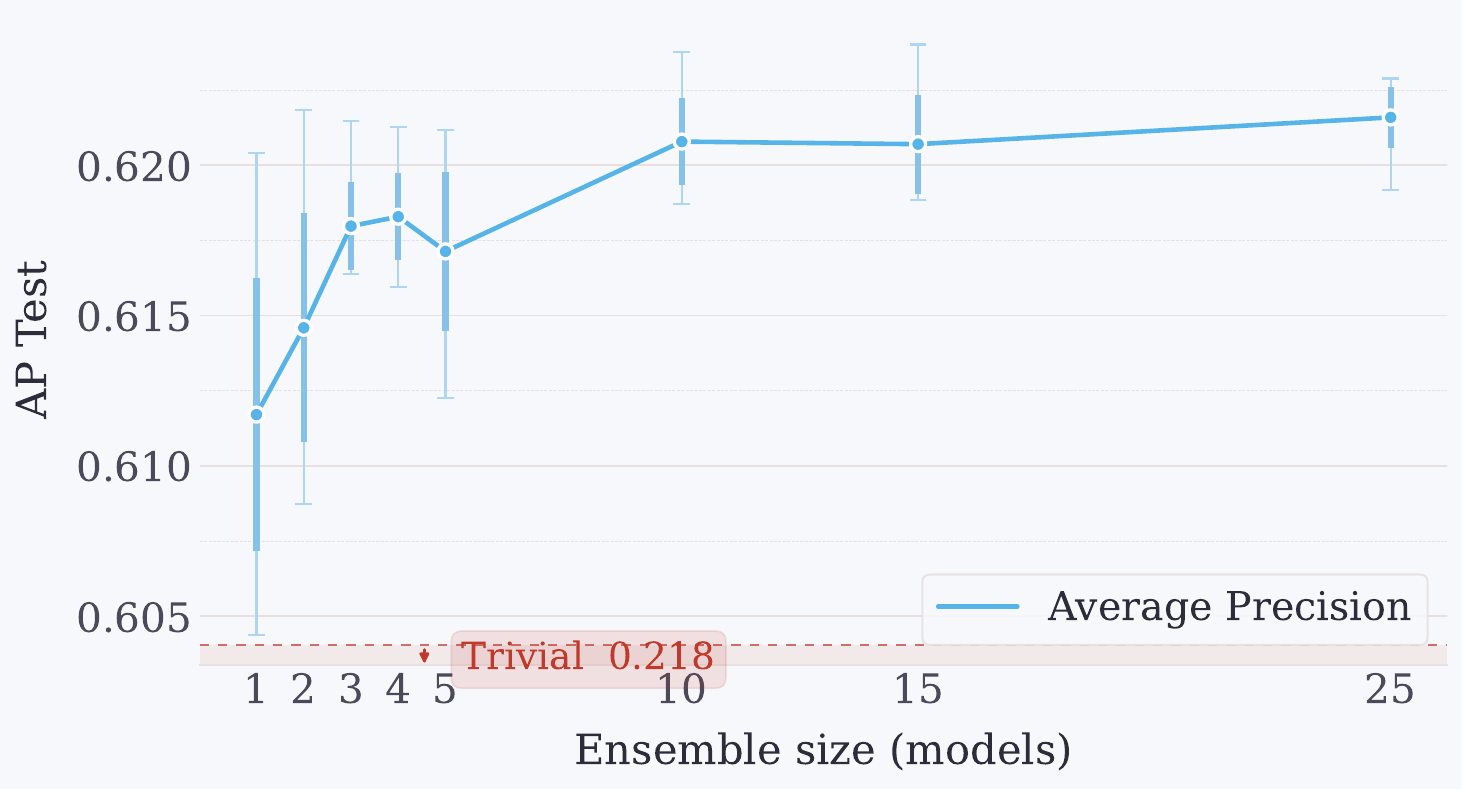}
        \caption{Tolokers2 Average Precision}
        \label{fig:tolokers2_point}
    \end{subfigure}

    \vspace{-0ex} 

    \begin{subfigure}{0.49\textwidth}
        \centering
        \includegraphics[width=\linewidth]{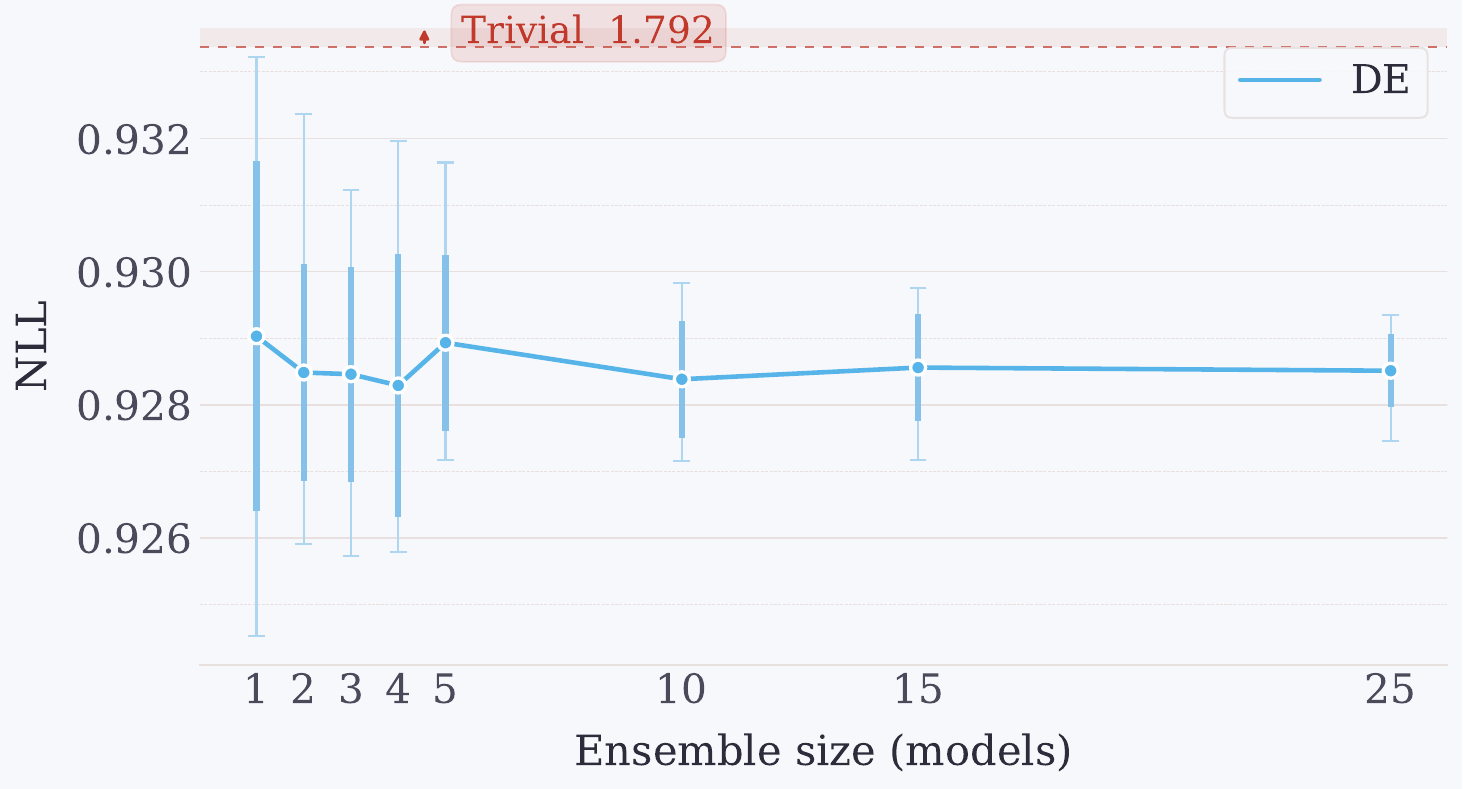}
        \caption{CiteSeer NLL}
        \label{fig:citeseer_nll}
    \end{subfigure}%
    \hfill
    \begin{subfigure}{0.49\textwidth}
        \centering
        \includegraphics[width=\linewidth]{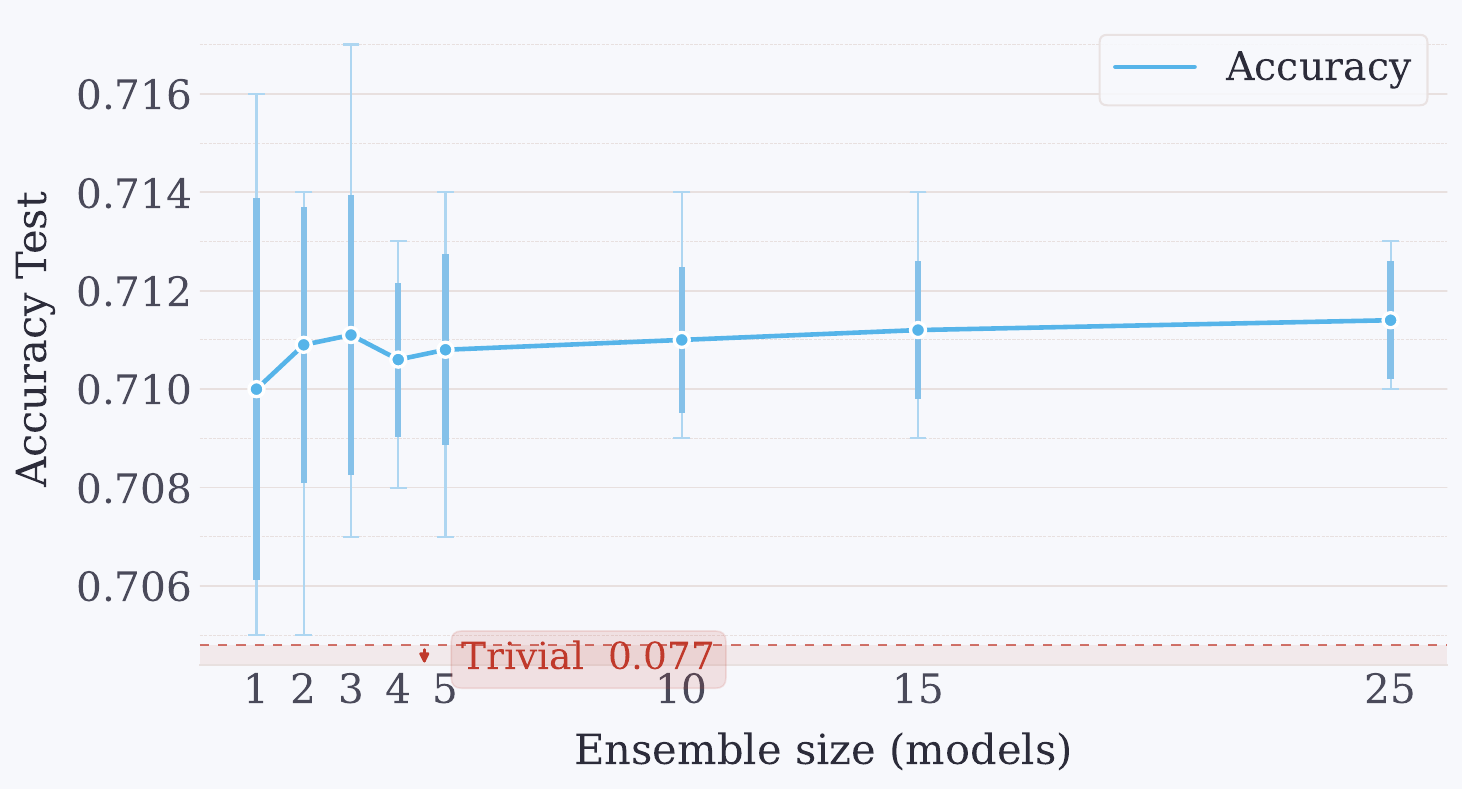}
        \caption{CiteSeer Accuracy}
        \label{fig:citeseer_point}
    \end{subfigure}

    \begin{subfigure}{0.49\textwidth}
        \centering
        \includegraphics[width=\linewidth]{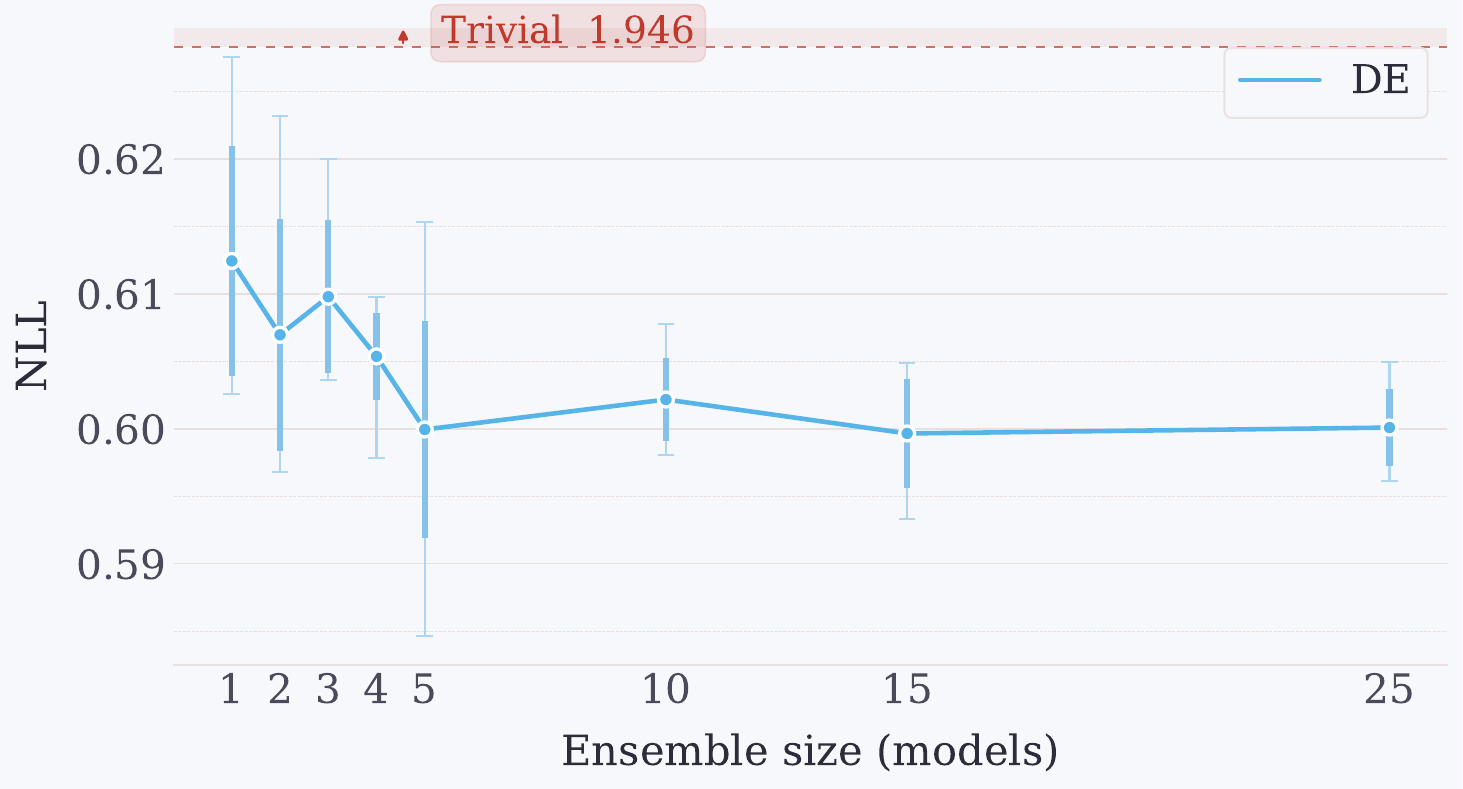}
        \caption{Cora NLL}
        \label{fig:cora_nll}
    \end{subfigure}%
    \hfill
    \begin{subfigure}{0.49\textwidth}
        \centering
        \includegraphics[width=\linewidth]{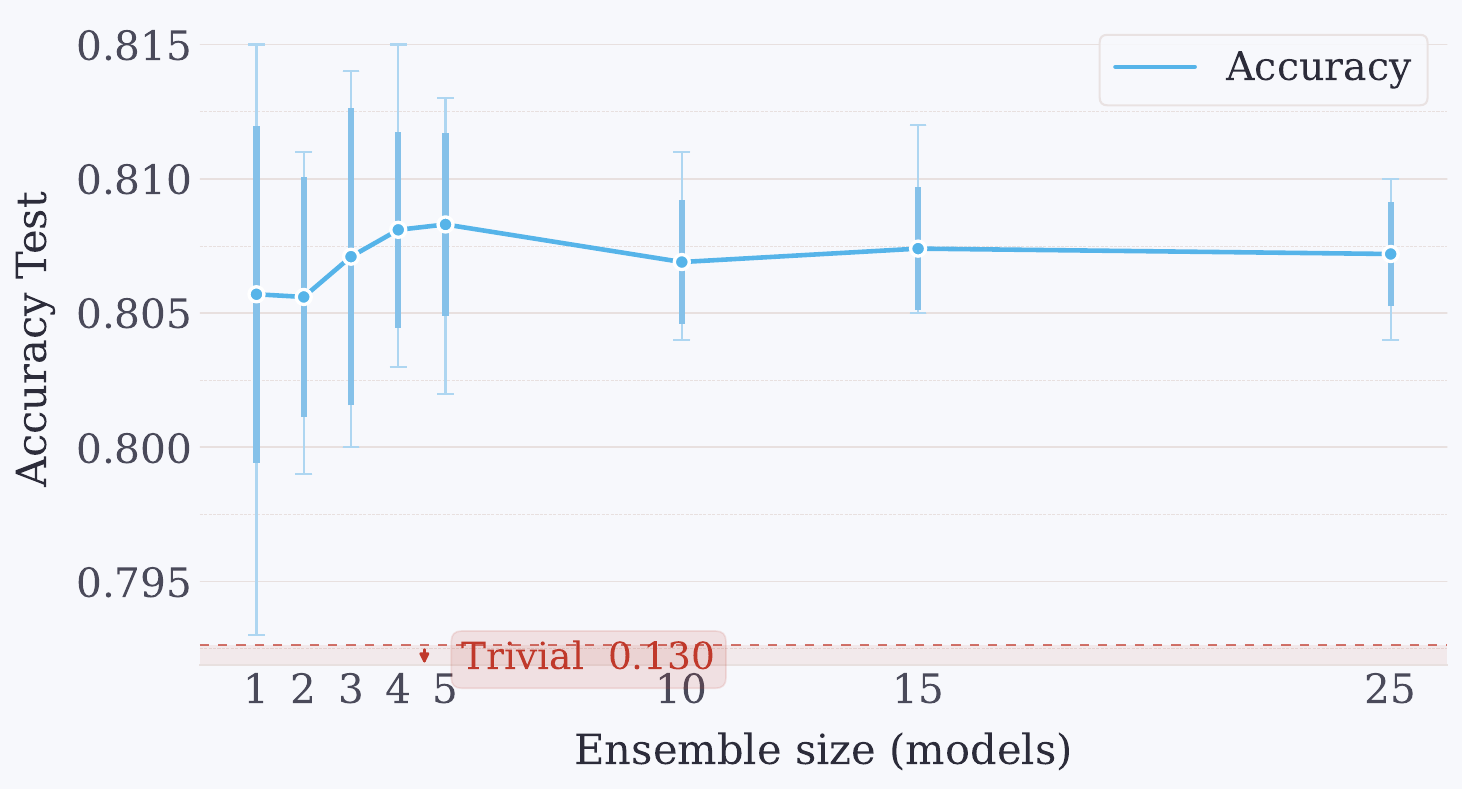}
        \caption{Cora Accuracy}
        \label{fig:cora_point}
    \end{subfigure}
    
    \caption{Evolution of NLL (\subref{fig:tolokers2_nll},~\subref{fig:citeseer_nll} and~\subref{fig:cora_nll}) and point estimation metric (\subref{fig:tolokers2_point},~\subref{fig:citeseer_point} and~\subref{fig:cora_point}) as the number of base models in the ensemble increases. Red dashed line denotes a trivial baseline.}
    \label{fig:compare_ensemble_size_classification:appendix}
\end{figure}

\begin{figure}[ht]
    \centering
    \begin{subfigure}{0.49\textwidth}
        \centering
        \includegraphics[width=\linewidth]{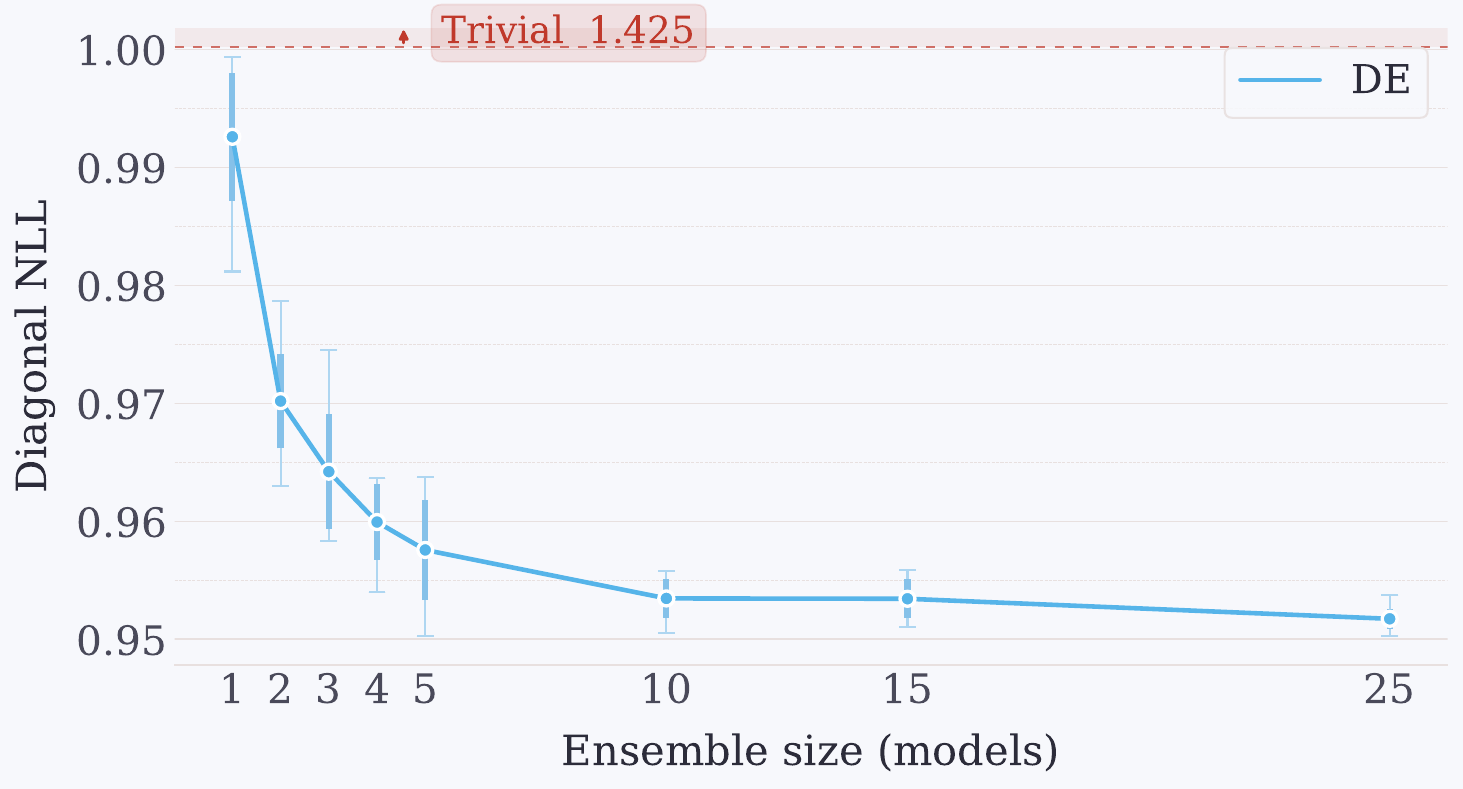}
        \caption{Artnetviews NLL}
        \label{fig:artnetviews_nll}
    \end{subfigure}%
    \hfill
    \begin{subfigure}{0.49\textwidth}
        \centering
        \includegraphics[width=\linewidth]{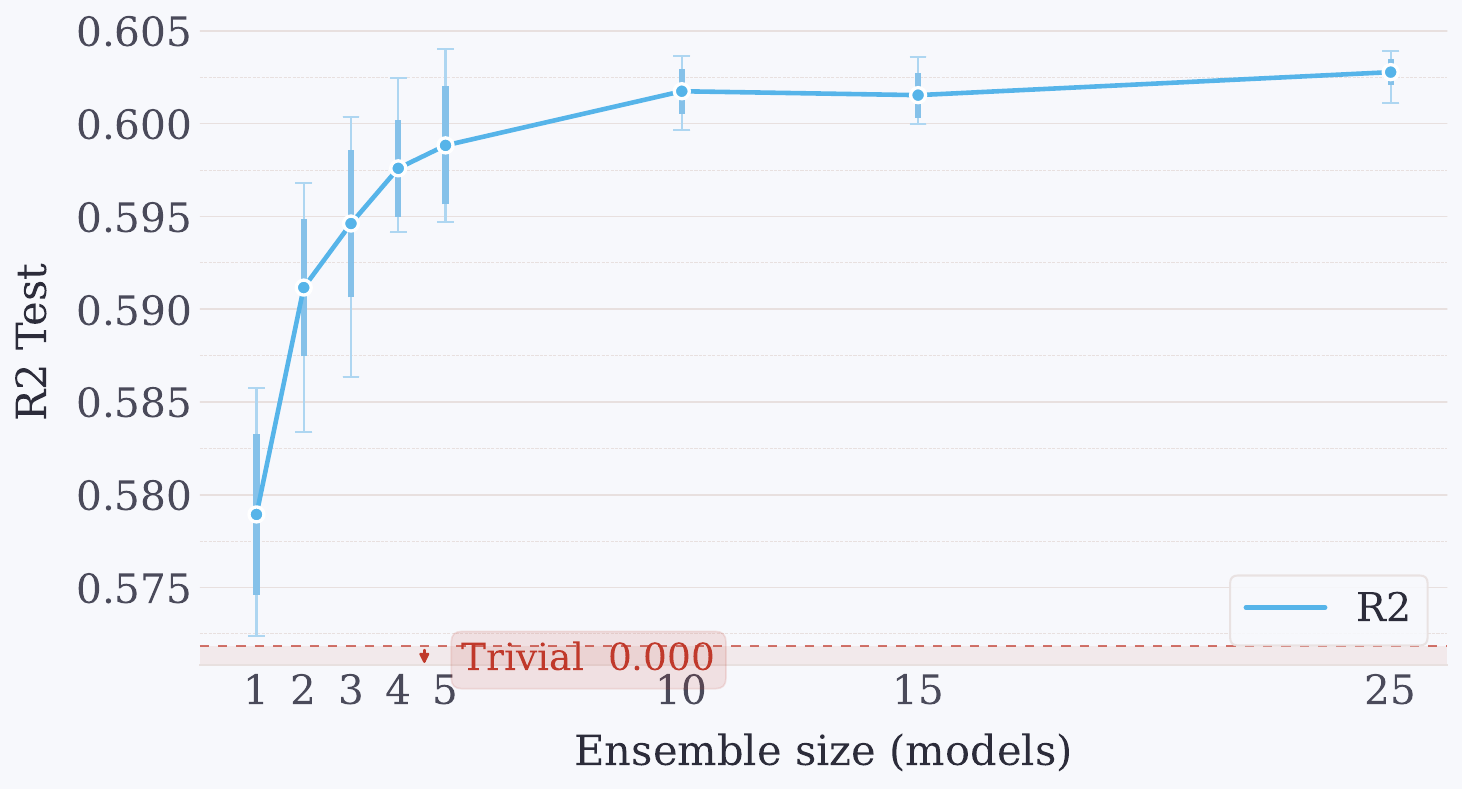}
        \caption{Artnetviews R2}
        \label{fig:artnetviews_point}
    \end{subfigure}

    \vspace{-0ex} 

    \begin{subfigure}{0.49\textwidth}
        \centering
        \includegraphics[width=\linewidth]{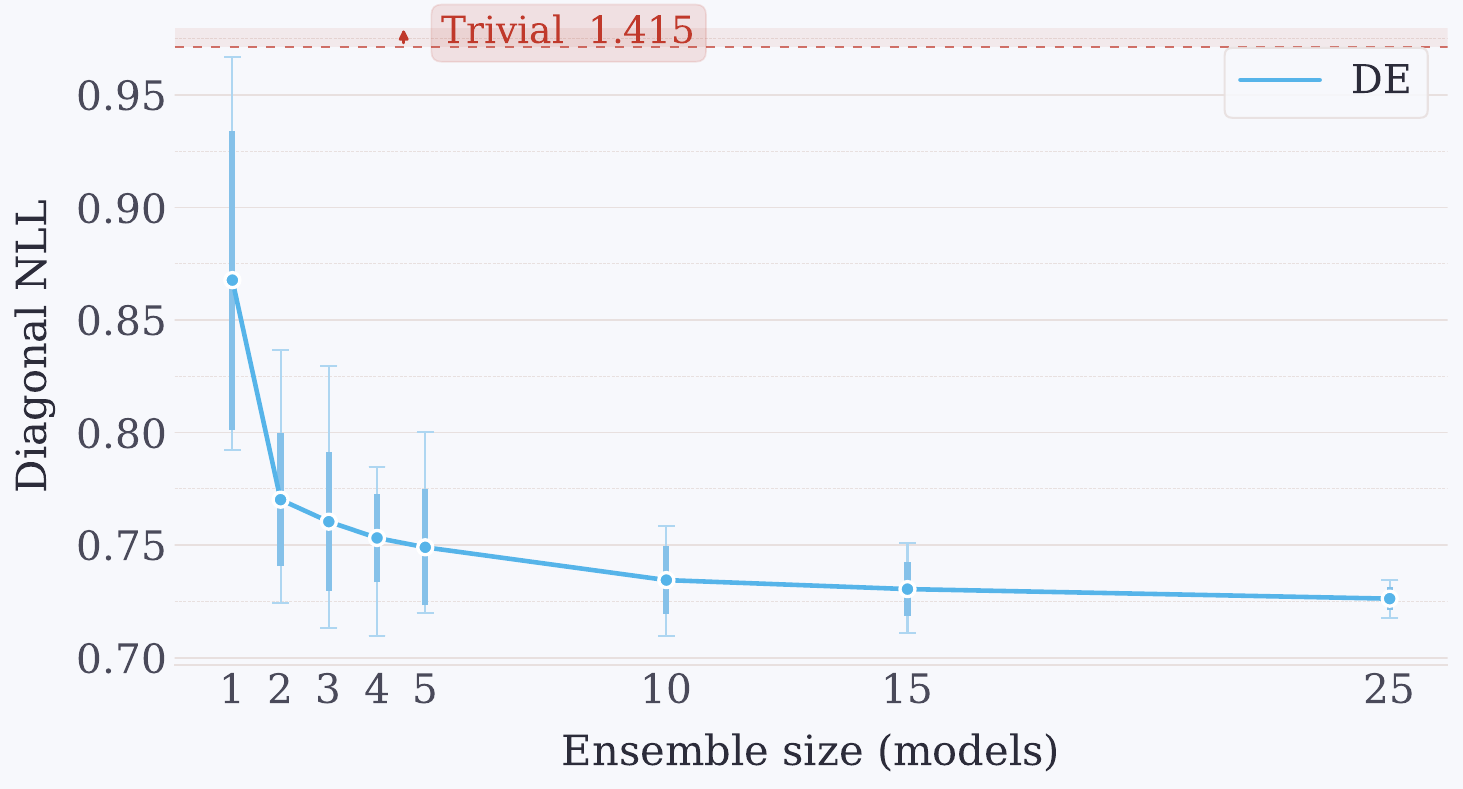}
        \caption{Chameleon NLL}
        \label{fig:chameleon_nll}
    \end{subfigure}%
    \hfill
    \begin{subfigure}{0.49\textwidth}
        \centering
        \includegraphics[width=\linewidth]{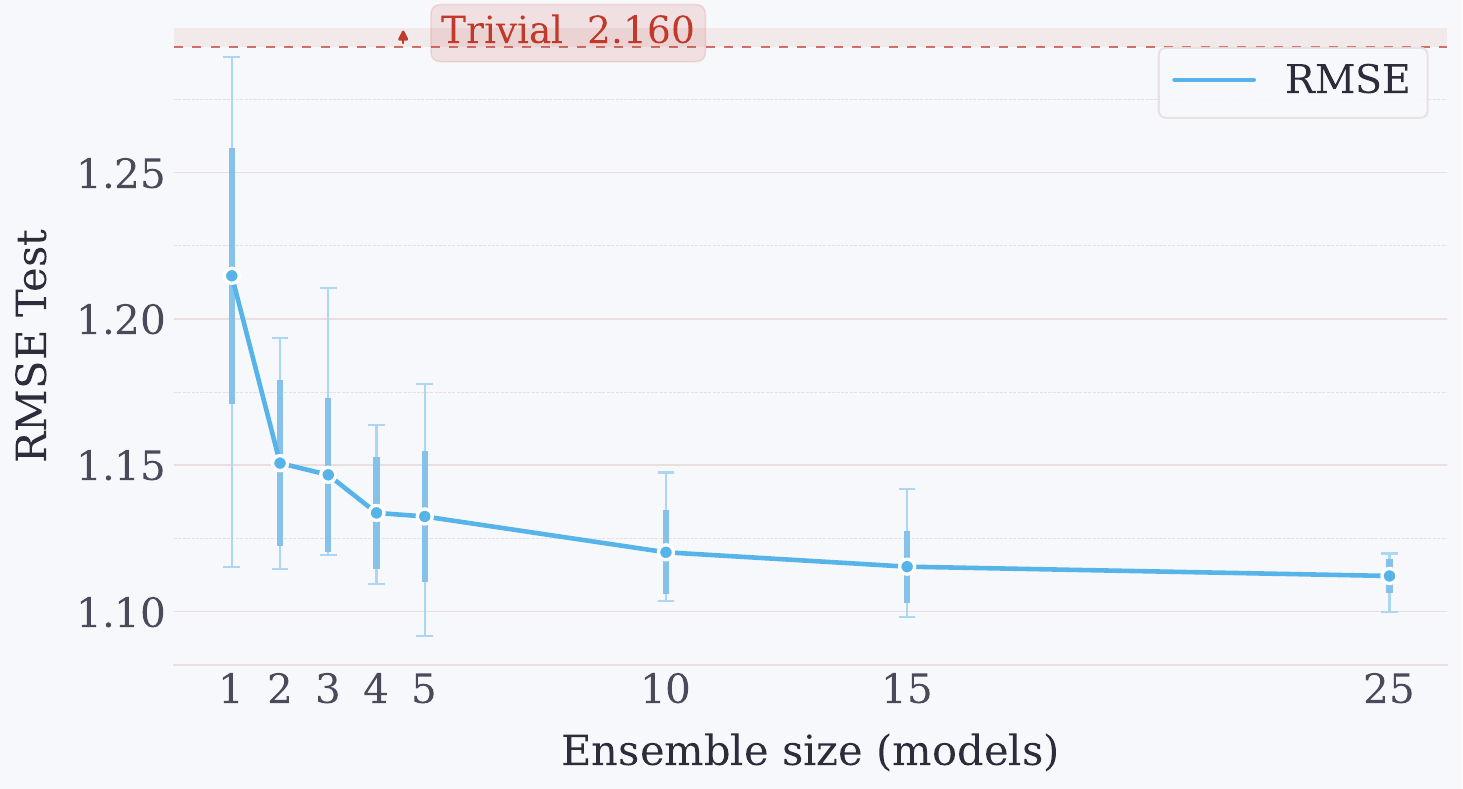}
        \caption{Chameleon RMSE}
        \label{fig:chameleon_point}
    \end{subfigure}

    \begin{subfigure}{0.49\textwidth}
        \centering
        \includegraphics[width=\linewidth]{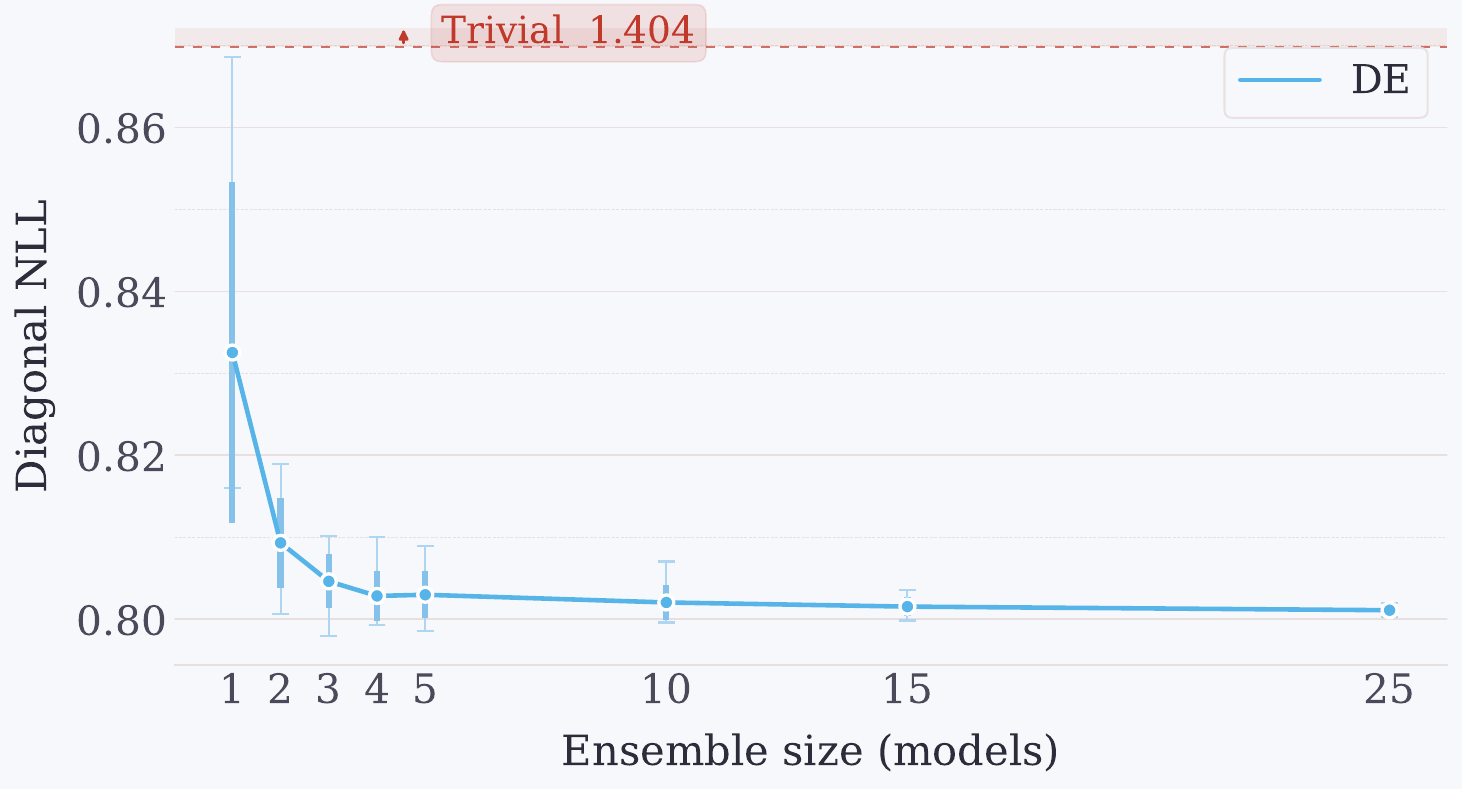}
        \caption{QM9 NLL}
        \label{fig:qm9_nll}
    \end{subfigure}%
    \hfill
    \begin{subfigure}{0.49\textwidth}
        \centering
        \includegraphics[width=\linewidth]{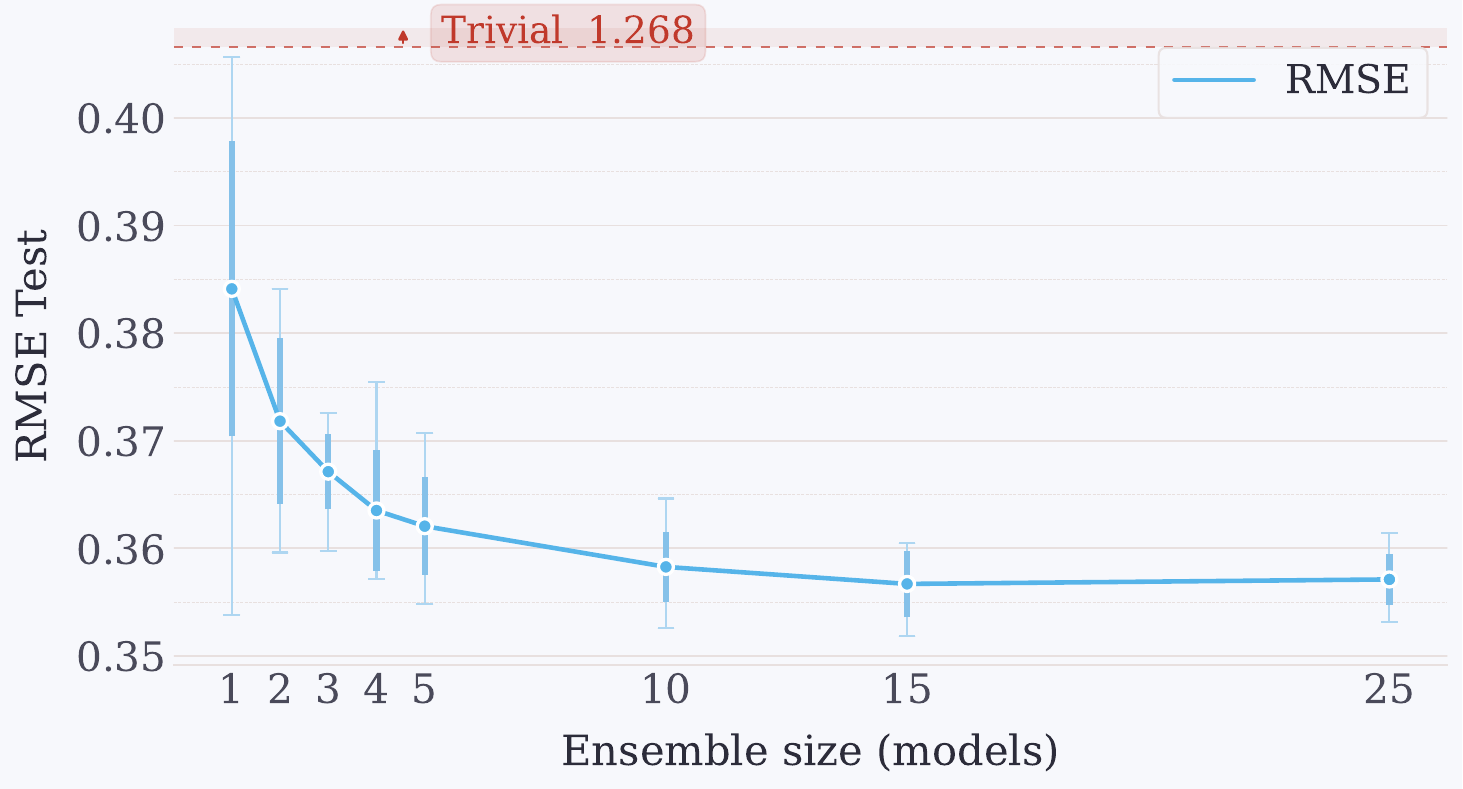}
        \caption{QM9 RMSE}
        \label{fig:qm9_point}
    \end{subfigure}

    \begin{subfigure}{0.49\textwidth}
        \centering
        \includegraphics[width=\linewidth]{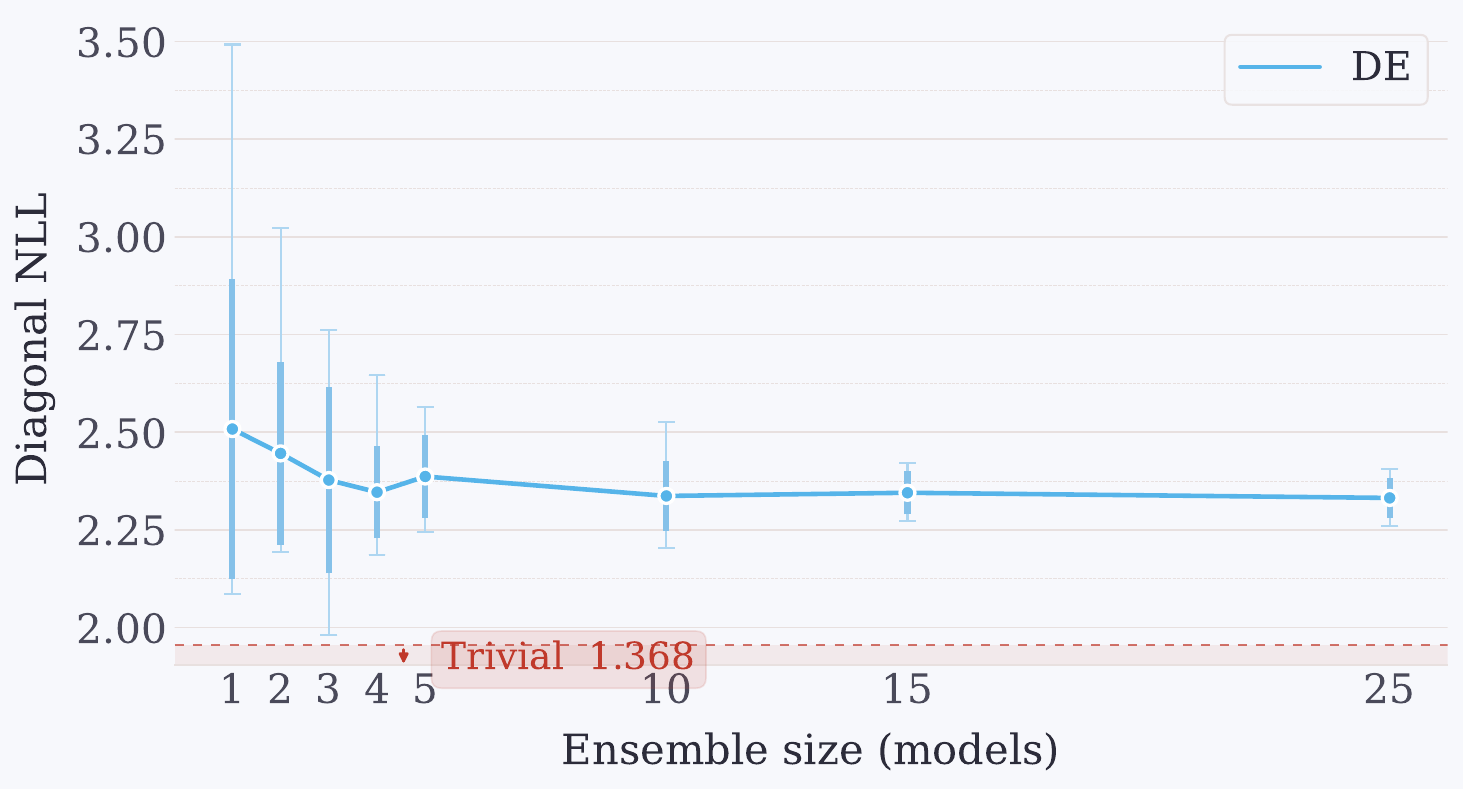}
        \caption{PEMS NLL}
        \label{fig:pems_nll}
    \end{subfigure}%
    \hfill
    \begin{subfigure}{0.49\textwidth}
        \centering
        \includegraphics[width=\linewidth]{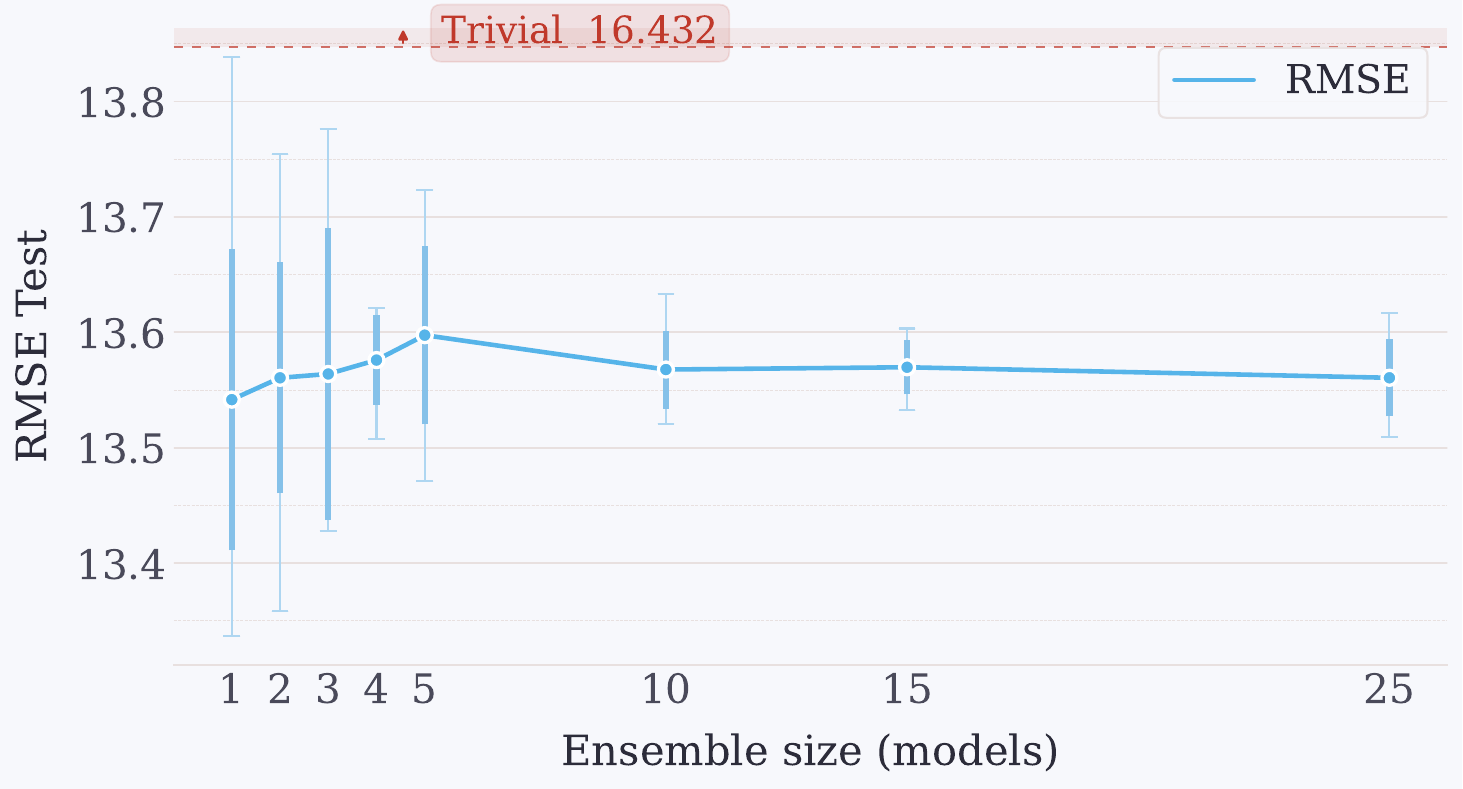}
        \caption{PEMS RMSE}
        \label{fig:pems_point}
    \end{subfigure}
    
    \caption{Evolution of NLL (\subref{fig:artnetviews_nll},~\subref{fig:chameleon_nll},~\subref{fig:qm9_nll} and~\subref{fig:pems_nll}) and point estimation metric (\subref{fig:artnetviews_point},~\subref{fig:chameleon_point},~\subref{fig:qm9_point} and~\subref{fig:pems_point}) as the number of base models in the ensemble increases. Red dashed line denotes a trivial baseline.}
    \label{fig:compare_ensemble_size_regression:appendix}
\end{figure}

\paragraph{Dynamics.}~\Cref{fig:compare_ensemble_size_regression:appendix} and~\Cref{fig:compare_ensemble_size_classification:appendix} show the dynamics of the point estimation metric and NLL for an increasing ensemble size.
For Artnetviews and Tolokers2 we revert back to the original metrics used in~\textcite{bazhenov2025graphland}.
This should allow one to more easily compare results if necessary.
Generally, the dynamics hold when changing metrics.
An added bonus is that average precision, compared with accuracy, makes the dynamics easier to see.

\vspace{1em}

\clearpage

\end{document}